\documentclass[letterpaper, 10 pt, conference]{ieeeconf}
\IEEEoverridecommandlockouts
\newcommand{\calG}{\mathcal{G}}
\newcommand{\calF}{\mathcal{F}}
\newcommand{\calM}{\mathcal{M}}
\newcommand{\calR}{\mathcal{R}}
\newcommand{\comment}[1]{}

\newtheorem{proposition}{Proposition}

\usepackage{graphics} 
\usepackage{epsfig} 
\usepackage{mathptmx} 
\usepackage{times} 
\usepackage{amsmath} 
\usepackage{amssymb}  

\title{\LARGE \bf Lagrangian Relaxation for MAP Estimation in
Graphical Models}

\author{Jason K. Johnson, Dmitry M. Malioutov and Alan
S. Willsky\thanks{The authors are with the Electrical Engineering and Computer Science Department, Massachusetts Institute of Technology, Cambridge,
MA 02139, USA.  {\tt\small {\{jasonj,dmm,willsky\}@mit.com}}.}}

\begin{document}

\maketitle
\begin{minipage}[t][0pt][t]{0.96\textwidth}
\vspace{-1.8in} 
In Proceedings of \emph{The 45th Allerton Conference on
Communication, Control and Computing}, September, 2007.
\end{minipage}\vspace{-\baselineskip}

\thispagestyle{empty}
\pagestyle{empty}
\begin{abstract}
We develop a general framework for MAP estimation in discrete and
Gaussian graphical models using Lagrangian relaxation techniques.  The
key idea is to reformulate an intractable estimation problem as one
defined on a more tractable graph, but subject to additional
constraints. Relaxing these constraints gives a tractable dual
problem, one defined by a thin graph, which is then optimized by an
iterative procedure.  When this iterative optimization leads to a
consistent estimate, one which also satisfies the constraints, then it
corresponds to an optimal MAP estimate of the original model.
Otherwise there is a ``duality gap'', and we obtain a bound on the
optimal solution.  Thus, our approach combines convex optimization
with dynamic programming techniques applicable for thin graphs. The
popular tree-reweighted max-product (TRMP) method may be seen as
solving a particular class of such relaxations, where the intractable
graph is relaxed to a set of spanning trees.  We also consider
relaxations to a set of small induced subgraphs, thin subgraphs
(e.g. loops), and a connected tree obtained by ``unwinding'' cycles.
In addition, we propose a new class of multiscale relaxations that
introduce ``summary'' variables.  The potential benefits of such
generalizations include: reducing or eliminating the ``duality gap''
in hard problems, reducing the number or Lagrange multipliers in the
dual problem, and accelerating convergence of the iterative
optimization procedure.
\end{abstract}

\section{Introduction}

Graphical models are probability models for a collection of random
variables on a graph: the nodes of the graph represent random
variables and the graph structure encodes conditional independence
relations among the variables.  Such models provide compact
representations of probability distributions, and have found many
practical applications in physics, statistical signal and image
processing, error-correcting coding and machine learning.  However,
performing optimal estimation in such models using standard junction
tree approaches generally is intractable in large-scale estimation
scenarios.  This motivates the development of variational techniques
to perform approximate inference, and, in some cases, recover the
optimal estimate.

We consider a general Lagrangian relaxation (LR) approach to
\emph{maximum a posteriori} (MAP) estimation in graphical models. The
general idea is to reformulate the estimation problem on an
intractable graph as a constrained estimation over an augmented model
defined on a larger, but more tractable graph.  Then, using Lagrange
multipliers to relax the constraints, we obtain a tractable estimation
problem that gives an upper-bound on the original problem.  This leads
to a convex optimization problem of minimizing the upper-bound as a
function of Lagrange multipliers.

We consider a variety of strategies to augment the original graph.
The simplest approach breaks the graph into many small, overlapping
subgraphs, which involves replicating some variables.  Similarly, the
graph can be broken into a set of thin subgraphs, as in the TRMP
approach, or ``unrolled'' to obtain a larger, but connected, thin
graph. We show that all of these approaches are essentially
equivalent, being characterized by the set of maximal cliques of the
augmented graph.  More generally, we also consider the introduction of
``summary'' variables, which leads naturally to multiscale algorithms.
We develop a general optimization approach based on marginal and
max-marginal matching procedures, which enforce consistency between
replicas of a node or edge, and moment-matching in the multiscale
relaxation. We show that the resulting bound is tight if and only if
there exists an optimal assignment in the augmented model that
satisfies the constraints.  In that case, we obtain the desired MAP
estimate of the original model.  When there is a duality gap, this is
evidenced by the occurrence of ``ties'' in the resulting set of
max-marginals, which requires further augmentation of the model to
reduce and ultimately eliminate the duality gap.  We focus primarily
on discrete graphical models with binary variables, but also consider
the extension to Gaussian graphical models.  In the Gaussian model, we
find that, whenever LR is ``well-posed'', so that the augmented model
is valid, it leads to a tight bound and the optimal MAP estimate, and
also gives \emph{upper-bounds} on variances that provide a measure of
confidence in the MAP estimate.

\section{Background}

We consider probabilistic graphical models
\cite{Lauritzen96,Cowell*99,Frey98}, which are probability
distributions of the form
\begin{equation}\label{eq:1}
p(x_1,\dots,x_n) = \frac{1}{Z} \exp\{f(x)\} = \frac{1}{Z} \exp\left\{\sum_{C \in \calG} f_C(x_C)\right\}
\end{equation}
where each function $f_C$ only depends on a subset of variables $x_C =
(x_v, v \in C)$ and $Z$ is a normalization constant of the model,
called the \emph{partition function} in statistical physics.  If the
sum ranges over all \emph{cliques} of the graph, which are the fully
connected subsets of variables, this representation is sufficient to
realize any Markov model on $\calG$ \cite{Lauritzen96}.
\comment{\footnote{The probability
distribution $p(x)$ is \emph{Markov} on $\calG$ if for every $S
\subset V$ that separates $A,B \subset V$ in $\calG$, $x_A$ and $x_B$
are independent given $x_S$.}}  However, it is also common to consider
restricted Markov models where only singleton and pairwise
interactions are specified.  In general, we specify the set of
interactions by a hypergraph $\calG \subset 2^V$, where $2^V$
represents the set of all subsets of $V$.  The elements of $\calG$
are its \emph{hyperedges}, which generalizes the usual concept of
a graph with pairwise edges.

\emph{Discrete Models.} While our approach is applicable for general
discrete models, we focus on models with binary variables.  One may
use either the Boltzmann machine representation $x_v \in \{0,1\}$, or
that of the Ising model $x_v \in \{-1,+1\}$.  These models can be
represented as in (\ref{eq:1}) with
\begin{equation}
f(x;\theta) = \sum_{E\in\calG} \theta_E \phi_E(x_E), \;\; \phi_E(x_E) = \prod_{v \in E} x_v
\end{equation}
This defines an \emph{exponential family} \cite{WainwrightJordan03} of
probability distributions based on model features $\phi$ and
parameterized by $\theta$.  $\Phi(\theta) \triangleq \log Z(\theta)$
is the \emph{log-partition function} and has the
\emph{moment-generating property}: $\frac{\partial
\Phi(\theta)}{\partial \theta_E} = \mathbb{E}_\theta\{\phi_E(x)\}
\triangleq \eta_E$.  Here, $\eta$ are the \emph{moments} of the
distribution, which serve both as an alternate parameterization of
the exponential family and, in graphical models, to specify the
marginal distributions on cliques of the model. Inference in discrete
models using junction tree methods, either to compute the mode or the
marginals, is generally linear in the number of variables $n$ but
grows exponentially in the \emph{width} of the graph \cite{Cowell*99}, which
is determined by the size of the maximal cliques in a junction tree
representation of the graph. Hence, exact inference is only tractable
for \emph{thin} graphs, that is, where one can build an equivalent
junction tree with small cliques.

\emph{Gaussian Models.} We also consider Gaussian graphical models
\cite{Dempster72,SpeedKiiveri86} represented in \emph{information form}:
\begin{equation}
p(x) = \exp\{-\tfrac{1}{2} x^T J x + h^T x - \Phi(h,J) \}
\end{equation}
where $J$ is the \emph{information matrix}, $h$ a potential vector and
$\Phi(h,J) = \tfrac{1}{2} \{ h^TJ^{-1}h - \log\det J + n\log 2\pi\}$.  This
corresponds to the standard form of the Gaussian model specified by
the covariance matrix $P = J^{-1}$ and mean vector $\hat{x} = J^{-1}
h$.  This translates into an exponential family where we identify
$(h,J)$ with the parameters $\theta$ and $(\hat{x},P)$ with the
moments $\eta$.  In general, the complexity of inference in Gaussian
models is $\mathcal{O}(n^3)$.  The fill pattern of $J$ determines the
Markov structure of the Gaussian model: $(i,j) \in \calG$ if $J_{i,j}
\neq 0$. Using more efficient recursive inference methods that exploit
sparsity, such as junction trees or sparse Gaussian elimination, the
complexity is linear in $n$ but cubic in the width of the graph, which
is still impractical for many large-scale estimation problems.

\section{Discrete Lagrangian Relaxation}

To begin with, consider the problem of maximizing the following
objective function, defined over a hypergraph $\calG \subset
2^V$ based on a vertex set $V = \{1,\dots,n\}$ corresponding to 
discrete variables $x = (x_1,\dots,x_n)$.
\begin{equation}
f(x) = \sum_{E \in \calG} f_E(x_E)
\end{equation}
For instance, this may be defined as $f(x) = \langle \theta, \phi(x)
\rangle$ in an exponential family graphical model, such that each term
corresponds to a feature $f_E(x_E) = \theta_E \phi_E(x_E)$.  Then, we
seek $x^*$ to maximize $f(x)$ to obtain the MAP estimate of (\ref{eq:1}).

\begin{figure}
\centering
\input{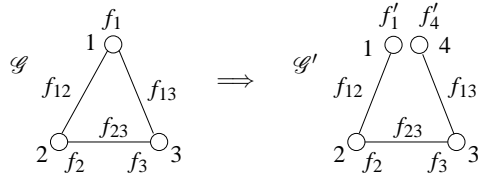}
\caption{\label{fig:toy_example}A simple illustrative example of Lagrangian relaxation.}
\vspace{-.3cm}
\end{figure}

\emph{An Illustrative Example.} To briefly convey the basic concept,
we consider a simple pairwise model defined on a $3$-node cycle
$\calG$ represented in Fig. \ref{fig:toy_example}. Here, the augmented
graph $\calG'$ is a 4-node chain, where node $4$ is a replica of node
$1$. We copy all the potentials on the nodes and edges from $\calG$ to
$\calG'$. For the replicated variables, $x_1'$ and $x_4'$, we split
$f_1$ between $f_1'$ and $f_4'$ such that $f_1(y) = f_1'(y) + f_4'(y)$
for $y \in \{0,1\}$. Now the problem $\max_x f(x)$ is equivalent to
maximizing $f'(x')$ subject to the constraint $x_1' = x_4'$. To solve
the latter we relax the constraint using Lagrange multipliers:
$L(x', \lambda) = f'(x') + \lambda(x'_1 - x'_4)$. The additional term
$\lambda(x_1' - x_4')$ modifies the self-potentials: $f_1' \leftarrow
f_1'(x_1') + \lambda x_1'$ and $f_4' \leftarrow f_4'(x_4') - \lambda
x_4'$, parameterizing a family of models on $\calG'$ all of which are
equivalent to $f$ under the constraint $x'_1=x'_4$. For a fixed
$\lambda$, solving $\max_x L(x, \lambda) \triangleq g(\lambda)$ gives
an upper bound on $f^* = \max_x f(x)$, so by optimizing $\lambda$ to
minimize $g(\lambda)$, we find the tightest bound $g^* = \min_\lambda
g(\lambda)$. If the constraint $x_1' = x_4'$ is satisfied in the final
solution, then there is strong duality $g^* = f^*$ and we obtain the
correct MAP assignment for $f(x)$.

We now discuss the general procedure and develop our approach to
optimize $g(\lambda)$ in more difficult cases.

\subsection{Obtaining a Tractable Graph by Vertex Replication} 

In this section, we consider approaches that involve
\emph{replicating} variables to define the augmented model. The basic
constraints in designing $\calG'$ are as follows: $\calG'$ is
comprised of replicas of nodes and edges of $\calG$.  Every node and
edge of $\calG$ must be represented at least once in $\calG'$.
Finally, $\calG'$ should be a thin graph, which relates to the
complexity of our method.

\begin{figure}
\centering
(a)\epsfig{file=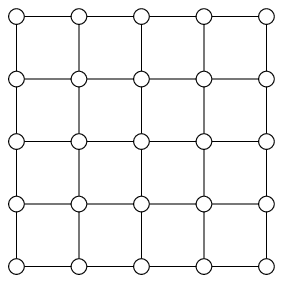,scale=.85}
\hspace{.2cm}
(b)\epsfig{file=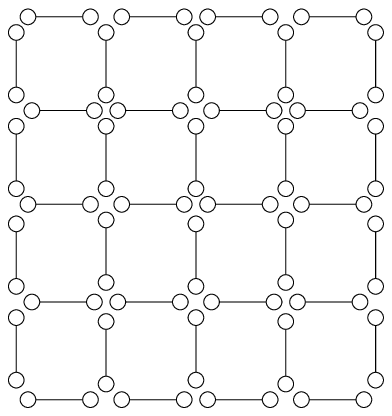,scale=0.55}\\
\vspace{.2cm}
(c)\epsfig{file=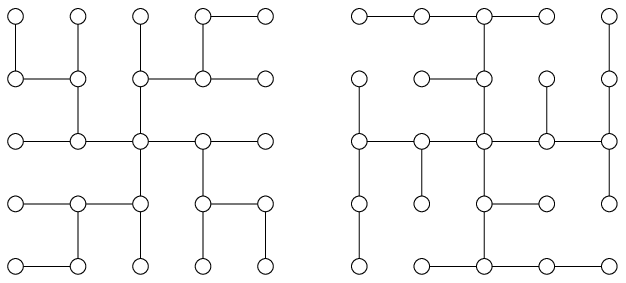,scale=0.7}
(d)\epsfig{file=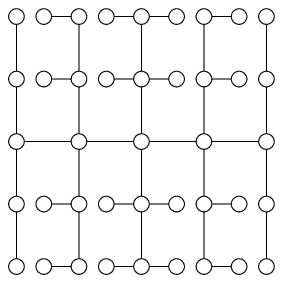,scale=0.7}\\
\vspace{.2cm}
(e)\epsfig{file=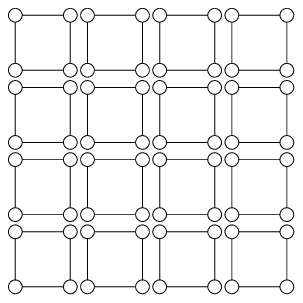,scale=0.75}
\hspace{.2cm}
(f)\epsfig{file=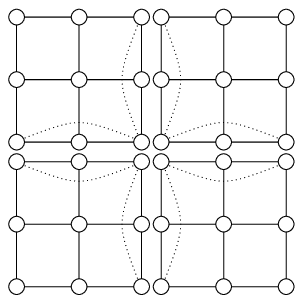,scale=0.75}\\
\vspace{.2cm}
(g)\epsfig{file=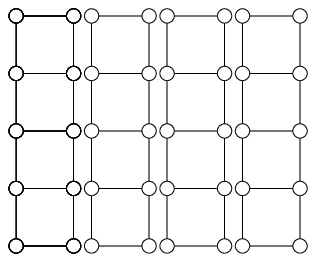,scale=0.75}
\hspace{.2cm}
(h)\epsfig{file=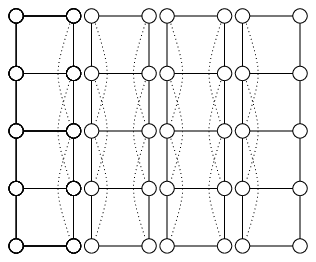,scale=0.75}
\caption{\label{fig:graphs} Illustrations of a variety of possible
ways to obtain a tractable graph structure from a $5 \times 5$ grid by
replicating some vertices of the graph.}
\vspace{-.4cm}
\end{figure}

To help illustrate the various strategies, we consider a
pairwise model $f(x)$ defined on $5 \times 5$ grid, as seen in
Fig.~\ref{fig:graphs}(a).  A natural approach is to break the
model up into small subgraphs.  The simplest method is to break the
graph up into its composite interactions.  For pairwise models, this
means that we split the graph into a set of disjoint edges as shown in
(b). Here, each internal node of the graph is replicated four times.
To reduce the number of replicated nodes, and hence the number of
constraints, it is also useful to merge many of these smaller
subgraphs into larger thin graphs.  One approach is to group edges
into \emph{spanning trees} of the graph as seen in (c).  Here, each
edge must be including in at least one tree, and some edges are
replicated in multiple trees.  The TRMP approach is based on this
idea.  One could also allow multiple replicas of a node in the same
connected component of $\calG'$.  For instance, by taking a spanning
tree of the graph and then adding an extra leaf node for each missing
edge we obtain the graph seen in (d).

It is also tractable to use small subgraphs that are not trees.  We
can break the graph into a set of short loops as in (e) or a set of
induced subgraphs as in (f) where we select a set of $3\times3$
subgraphs that overlap on their boundary.  In such cases, including
additional edges in the overlap of these subgraphs, such as the dotted
edges in (f), can enhance the relaxations that we consider.  Finally,
we reduce the number of constraints in these formulations by again
grouping subgraphs to form larger subgraphs that are still thin, as
shown in (g).  This will also lead to tractability in our methods.
Again, it can be useful to include extra edges in the overlap of these
subgraphs as in (h), although this increases the width of the subgraph
and affects the computational complexity of our methods.

\emph{Notation.} Let $\calG^\prime$ denote the augmented graph (or
collection of subgraphs), which is based on an extended vertex set
$V'$, comprised of replicas of nodes in $V$. We assume that all edges
of this graph are also replicas of edges of the original graph
$\calG$.\footnote{In the case that we introduce extra edges in
$\calG'$, as in (f) and (h), we also add corresponding edges to
$\calG$ to maintain this convention.} Thus, there is a
well-defined surjective map $\Gamma: \calG' \rightarrow \calG$, each
edge $E' \in \calG'$ is a replica an edge $E=\Gamma(E') \in
\calG$, and every edge of $\calG$ has at least one such replica.  This
notation is overloaded for nodes by treating them as singleton edges
of $\calG$.  We also denote the set-valued inverse of $\Gamma$ by
$\mathcal{R}(E) \triangleq \Gamma^{-1}(E)$, which is the set of
replicas of $E$, and let $r_E \triangleq |\calR(E)|$ denote the
number of replicas.  This defines an equivalence relation on $\calG'$:
$A,B \in \calG'$ are equivalent $A \equiv B$ if $\Gamma(A)=\Gamma(B)$,
that is, if $A,B \in \calR(E)$ are replicas of the same edge $E \in
\calG$.

\subsection{Equivalent Constrained Estimation Problem}

We now define a corresponding objective function $f'(x')$, where $x' =
(x'_v)_{v \in V'}$ are the variables of the augmented model. For each
hyperedge $E \in \calG$ (including individual nodes), we split the
function $f_E(x_E)$ among a set of replica functions $\{f'_{E'}, E'
\in \calR(E)\}$, requiring that these are \emph{consistent},
\begin{equation}
f_E(x_E) = \sum_{E' \in \calR(E)} f'_{E'}(x_E) \mbox{ for all } x_E.
\end{equation}
Using the parametric representation $f(x) = \langle \theta, \phi(x)
\rangle$, this consistency condition is equivalent to requiring
$\theta_E = \Sigma_{E'} \theta'_{E'}$. We will see that the LR
approach to follow may be viewed as an optimization over all such
possible consistent splittings.  Next, we define the augmented
objective function over the graph $\calG'$ as
\begin{equation}
f'(x') \triangleq \sum_{E \in \calG'} f'_E(x'_E).
\end{equation}
This insures that $f(x) = f'(x')$ where $x' = \zeta(x)$ is the
replicated version of $x$, defined by $x'_{v'} = x_v$ for all $v' \in
\calR(v)$.  This equivalence holds for all \emph{consistent}
configurations $x^\prime \in \zeta(\mathbb{X})$, where $x'$ is
self-consistent over various replicas of the same node.  Thus, we are
led to an equivalent optimization problem in the augmented model
subject to consistency constraints:
\begin{equation}
\label{eq:a}
f^* \triangleq \max_{x \in \mathbb{X}} f(x) = \max_{x' \in
\zeta(\mathbb{X})} f'(x')
\end{equation}
Expressing the consistency constraint as a set of linear constraints
on the model features $\phi$, we obtain:
\begin{equation}
\begin{array}{ll}
\mbox{maximize} & f'(x')\\
\mbox{subject to} & \phi_A(x'_A) = \phi_B(x'_B) \mbox{ for all } A \equiv B.
\end{array}
\end{equation}
Recall that, in the discrete binary model, these features are defined
$\phi_E(x_E) = \Pi_{v \in E} \, x_v$. Clearly, there is some
redundancy in these constraints: $x_a = x_b$ for all replicated nodes
$a \equiv b$ would insure that the edges agree. However, these
redundant edge-wise feature constraints do enhance the
following relaxation.

\subsection{Lagrangian Relaxation}

We have now defined an equivalent model on a tractable graph.
However, the equivalent \emph{constrained} optimization is still
intractable, because the constraints couple some variables of
$\calG^\prime$, spoiling its tractable structure.  This suggests the
use of Lagrangian duality to relax those complicating constraints.
Introducing Lagrange multipliers $\lambda_{A,B}$ for each constraint,
we define the \emph{Lagrangian}, which is a modified version of the
objective function:
\begin{equation}\label{eq:Lagrangian}
L(x',\lambda) = f'(x') + \sum_{A \equiv B} \lambda_{A,B} \,
(\phi_A(x'_A)-\phi_B(x'_B))
\end{equation}
Grouping terms by edges $E \in \calG'$, and using $f'_E(x_E) = \theta'_E \phi_E(x_E)$, this is represented
\begin{equation}
L(x',\lambda) = \sum_{E \in \calG'} f'_E(x'_E;\lambda) \nonumber \\
\end{equation}
\begin{equation}
f'_E(x'_E;\lambda) = \theta'_E(\lambda) \phi_E(x'_E) \nonumber \\
\end{equation}
\begin{equation}
\theta'_E(\lambda) = \theta'_E + \Sigma_B \lambda_{E,B} - \Sigma_A
\lambda_{A,E}
\end{equation}
 Note that the Lagrange multipliers may be interpreted as
parameterizing all consistent splittings, $\theta'(\lambda)$ spans the
subspace of all consistent $\theta'$ parameters.\footnote{We obtain a
minimal $\lambda$ parameterization by only using a subset of
constraints in (\ref{eq:Lagrangian}), such that
$\{(\phi_A(x')-\phi_B(x'))\}$ are linearly independent.}

It is tractable to maximize the Lagrangian, as it is defined
over the thin graph $\calG^\prime$. The value of this maximization
defines the \emph{dual function}:
\begin{equation}
g(\lambda) = \max_{x'} L(x',\lambda)
\end{equation}
Note that this is an \emph{unconstrained} optimization over
$\mathbb{X}^\prime$, and its solution need not lead to a consistent
$x' \in \zeta(\mathbb{X})$.  However, if this $x'$ is consistent then
it is an optimal solution of the constrained optimization problem
(\ref{eq:a}), and hence $x = \zeta^{-1}(x')$ (which is well-defined
for consistent $x'$) is also an optimal solution of the original
problem.  This is the goal of our approach, to find tractable
relaxations of the MAP estimation problem which lead to the correct
MAP estimate.  This motivates solution of the \emph{dual problem}:
\begin{equation}\label{eq:dual_problem}
\min_\lambda g(\lambda) \triangleq g^*
\end{equation}
Appealing to well-known results
\cite{Bertsekas95,BertsimasTsitsiklis97}, we conclude:

\begin{proposition}[Lagrangian duality] We have $g(\lambda) \ge f^*$ for all $\lambda$. Hence $g^* \ge f^*$.  If $g(\lambda^*)=g^*$, then one of the
following holds:
\begin{enumerate}
\item[(i)] There exists a consistent solution: 
\begin{displaymath}
x' \in \arg\max_{x' \in \mathbb{X}'} L(x';\lambda^*) \cap \zeta(\mathbb{X}).
\end{displaymath} 
Then, we have \emph{strong duality} $g^* = f^*$ and the set of \emph{all} MAP estimates is obtained as:
\begin{displaymath}
\arg\max_{x' \in \zeta(\mathbb{X})} f'(x') = \arg\max_{x' \in \mathbb{X}} L(x',\lambda^*) \cap \zeta(\mathbb{X}).
\end{displaymath}
\item[(ii)] There are no consistent solutions: 
\begin{displaymath}
\arg\max_{x' \in \mathbb{X}'} L(x';\lambda^*) \cap \zeta(\mathbb{X}) = \emptyset.
\end{displaymath} 
Then, there is a \emph{duality gap} $g^* > f^*$ and \emph{no} choice
of $\lambda$ will provide a consistent solution.
\end{enumerate}
Also, condition (i) holds only if $g(\lambda^*)=g^*$.
\end{proposition}

\begin{figure}
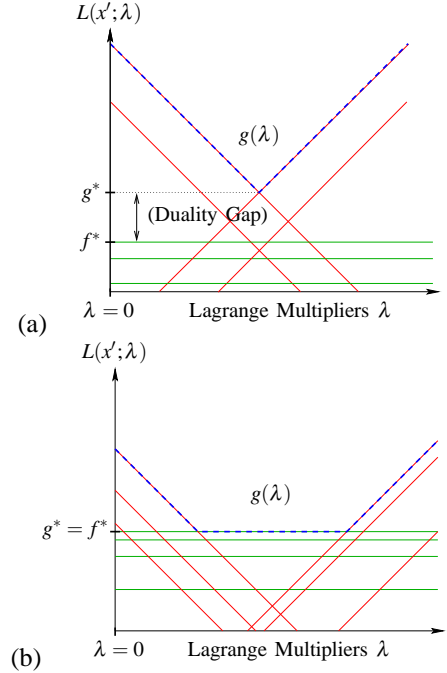

\centering 
(a)\input{dual_gap.pstex_t}
(b)\input{dual_func.pstex_t}
\caption{\label{fig:LR}Illustration of the Lagrangian duality in
the cases that (a) there is a duality gap and (b) there is no duality
gap (strong duality holds).}
\vspace{-.4cm}
\end{figure}

This result generalizes the analogous \emph{strong tree-agreement}
optimality condition for TRMP, and clarifies its connection to
standard Lagrangian duality results for integer programs.  To provide
some intuition, we present the following geometric interpretation
illustrated in Fig. \ref{fig:LR}.  The dual function is the maximum
over a finite set of linear functions in $\lambda$ indexed by $x'$.
For each $x' \in \mathbb{X}'$, there is a linear function
$g(\lambda;x') = \langle a(x'), \lambda\rangle + b(x')$, with $a(x') =
(\phi_A(x')-\phi_B(x'))_{A \equiv B}$, which is the gradient, and
$b(x') = f'(x')$.  The graph of each of these functions defines a
hyperplane in $\mathbb{R}^{d+1}$, where $d$ is the number of
constraints.  The flat hyperplanes, with $a = 0$, correspond to
consistent assignments $x' \in \zeta(\mathbb{X})$.  The remaining
sloped hyperplanes represent inconsistent assignments.  Hence, the
highest flat hyperplane corresponds to the optimal MAP estimate, with
height equal to $f^*$.  The dual function $g(\lambda)$ is defined by
the maximum height over this set of hyperplanes for each $\lambda$,
and is therefore convex, piece-wise linear and greater than or equal
to $f^*$ for all $\lambda$.  In the case of a duality gap, the
inconsistent hyperplanes hide the consistent ones, as depicted in (a),
so that the minimum of the dual function is defined by an intersection
of slanted hyperplanes corresponding to inconsistent assignments of
$x'$.  If there is no duality gap, as depicted in (b), then the
minimum is defined by the flat hyperplane corresponding to a
consistent assignment. Its intersection with slanted hyperplanes
defines the polytope of optimal Lagrange multipliers over which the
maximum flat hyperplane is exposed.

\subsection{Linear Programming Formulations}

We briefly consider a connection between this LR
picture and TRMP \cite{Wainwright*nov05,KolmogorovWainwright05} and
related linear programming approaches
\cite{Feldman*05,Yanover*06,Werner07}. This analysis also serves to
understand when different relaxations of the MAP estimation
problem will be equivalent.

The \emph{epigraph} of the dual function is defined as the set of all
points $(\lambda,h) = \mathbb{R}^{d+1}$ where $g(\lambda) \le h$, that
is, where $a(x') \lambda + b(x') \le h$ for all $x'$.  Thus, the
minimum of the dual function is equal to the lowest point of the
epigraph, which defines a linear program (LP) over $(\lambda,h) \in
\mathbb{R}^{d+1}$:
\begin{equation}
\begin{array}{ll}
\mbox{minimize} & h \\
\mbox{subject to} & \langle a(x'), \lambda \rangle + b(x') \le h \mbox{ for all } x'.
\end{array}
\end{equation}
Note that there are exponentially many constraints in this
formulation, so it is intractable.  However, recalling that it
\emph{is} tractable to compute the dual function for a given
$\lambda$, using the max-product algorithm applied to the thin graph
$\calG'$, we seek a more tractable representation of this LP.  To
achieve this, we consider the LP dual problem obtained by dualizing
the constraints, which is always tight \cite{BertsimasTsitsiklis97}.
This LP dual should be distinguished from our Lagrangian dual
(\ref{eq:dual_problem}) that is the subject of our paper.

Introducing non-negative Lagrange multipliers $\mu(x') \ge 0$ for each
inequality constraint, indexed by $x' \in \mathbb{X}'$, we obtain the 
LP Lagrangian:
\begin{eqnarray}
M(h,\lambda;\mu) &=& h + \mu\left[ \langle a(x'), \lambda \rangle + b(x') - h \right] \nonumber \\
 &=& \langle \mu[a], \lambda \rangle + \mu[b] + (1-\mu[1]) h,
\end{eqnarray} 
where $\mu$ denotes $\mu$-weighted summation, e.g.,
$\mu[a] = \sum_{x'} \mu(x') a(x')$.  The LP dual function is then:
\begin{equation}
M^*(\mu) \triangleq \min_{h,\lambda} M(h,\lambda;\mu) =
\left\{
\begin{array}{ll}
\mu[b], & \mu[1]=1 \mbox{ and } \mu[a]=0\\
-\infty, & \mbox{otherwise.}
\end{array}
\right.
\end{equation}
Note that $\mu > 0$ and $\mu[1]=1$ imply that $\mu$ is a probability
distribution and $\mu[\cdot]$ an expectation operator.  Recalling
$a(x') \triangleq (\phi_A(x')-\phi_B(x'), A \equiv B)$ and $b(x')
\triangleq f'(x')$, we obtain the dual LP:
\begin{equation}
\label{eq:b}
\max_{\mu \ge 0} M^*(\mu) =
\left\{
\begin{array}{ll}
\mbox{maximize} & \mu[f'] \\
\mbox{subject to} & \mu[\phi_A] = \mu[\phi_B] \mbox{ for } A \equiv B
\end{array}
\right.
\end{equation}
We seek a probability distribution over all configurations of the
augmented model that maximizes the expected value of $f'(x')$ subject
to constraints that the moments specifying marginal distributions
are consistent for replicated nodes and edges of the graph. This is a
convex relaxation of the constrained version of problem (4), where the
objective and constraint functions have been replaced by their
expected values under $\mu$.  Note that only marginals $\mu_{E'}$ over
hyperedges $E \in \calG'$ are needed to evaluate both the objective
and the constraints of this LP.  Hence, it reduces to one defined over
the \emph{marginal polytope} $\calM(\calG')$ \cite{Wainwright*nov05},
defined as the set of all \emph{realizable} collections of marginals
over the hyperedges of $\calG'$. Moreover, if the graph $\calG'$ is
\emph{chordal} \cite{Cowell*99}, then its marginal polytope has a
simple characterization. Let $\calM_{\mathrm{local}}(\calG')$ denote
the \emph{local marginal polytope} defined as the set of all edge-wise
marginal specifications that are consistent on intersections of edges.
In general, $\calM(\calG') \subset \calM_{\mathrm{local}}(\calG')$.
However, in chordal graphs it holds that $\calM(\calG') =
\calM_{\mathrm{local}}(\calG')$.  Thus, if $\calG'$ is a thin chordal
graph, we obtain a tractable LP whose value is equivalent to $g^*$ in
our framework.\footnote{Some graphs shown in Fig. \ref{fig:graphs} are
not chordal, but they can be extended to a thin chordal graph by
adding a few edges.  If no two of these new edges are equivalent when
mapped into $\calG$, then this does not change $g^*$.}

One last step shows the connection to LP approaches
\cite{Wainwright*nov05,Feldman*05,Yanover*06}.  The key observation is
that, roughly speaking,
\begin{equation}
\calM_{\mathrm{local}}(\calG') \cap \{\mu | \mu(x_A) = \mu(x_B), A \equiv B\} \equiv \calM_{\mathrm{local}}(\calG).
\end{equation}
This is seen by replicating marginals from $\calG$ to $\calG'$, or by
copying (consistent) replicated marginals back to $\calG$.
For such consistent $\mu$, we have $\mu[f'] = \mu[f]$, which gives:
\begin{equation}
g^* = \max_{\mu \in \calM_{\mathrm{local}}(\calG)} \mu[f] \ge \max_{\mu \in \calM(\calG)} \mu[f] = f^*.
\end{equation}
The maximum over $\calM_{\mathrm{local}}(\calG)$ gives an upper-bound on the maximum over $\calM(\calG) \subset \calM_{\mathrm{local}}(\calG)$. The latter is equivalent to exact MAP estimation and the bound becomes tight if $\calG$ is the set of maximal cliques of a chordal graph.  This discussion leads to the following characterization of LR:

\begin{proposition}[LR Hierarchy]
\emph{Equivalence:} Let $\calG'_1$ and $\calG'_2$ be the set of
maximal cliques of two chordal augmented graphs. If
$\Gamma^{-1}(\calG_1)=\Gamma^{-1}(\calG_2)$ then $g_1^*=g_2^*$ for the
respective dual problems.  Let $g^*(\calG)$ denote the common dual
value of all such chordal relaxations where
$\Gamma^{-1}(\calG')=\calG$. \emph{Monotonicity:} If $\calG_1 \subset
\calG_2$ then $g^*(\calG_1) \ge g^*(\calG_2)$.  \emph{Strong Duality:}
If $\calG$ is the set of maximal cliques of a chordal graph, then
$g^*(\calG)=f^*$.
\end{proposition}

\subsection{Smooth Relaxation of the Dual Problem}

\begin{figure}
\centering
\input{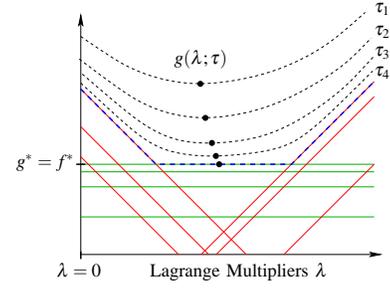}
\caption{\label{fig:smoothLR} Illustration of the ``log-sum-exp''
smooth approximation of the dual function, as a function of
``temperature'' $\tau$, and of an optimization procedure for
minimizing the non-smooth dual function through a sequence of smooth
minimizations.}
\vspace{-.4cm}
\end{figure}

In this section, we develop an approach to solve the dual problem.
One approach to minimize $g(\lambda)$ is to use non-smooth
optimization methods, such as the subgradient method
\cite{BertsimasTsitsiklis97}.  Here, we consider an
alternative, based on the following smooth approximation of
$g(\lambda)$:
\begin{equation}
g(\lambda; \tau) \triangleq \tau \log \sum_{x' \in \mathbb{X}} \exp\left( \frac{L(x';\lambda)}{\tau}\right)
\end{equation}
As illustrated if Fig. \ref{fig:smoothLR}, the parameter $\tau > 0$
controls the trade-off between smoothness of $g(\lambda;\tau)$ and how
well it approximates $g(\lambda)$.  This is known as the
``log-sum-exp'' approximation to the ``max function''
\cite{BoydVandenberghe04}: 
\begin{equation}
g(\lambda) \le g(\lambda;\tau) \le g(\lambda) + \tau \log |\mathbb{X}| \mbox{ for all } \tau > 0.
\end{equation}  
Hence, $g(\lambda;\tau) \rightarrow g(\lambda)$ \emph{uniformly} as $\tau
\rightarrow 0$ and, hence, $g^*(\tau) \triangleq \min_\lambda g(\lambda;\tau)$
converges to $g^*$.

The function $g(\lambda;\tau)$ has another useful interpretation.  Consider
the Gibbs distribution defined by
\begin{equation}
p_{\lambda,\tau}(x') = \exp\left( \frac{L(x',\lambda) - g(\lambda;\tau)}{\tau} \right)
\end{equation}
Here, $\tau > 0$ is the ``temperature'' and $g(\lambda;\tau)$
normalizes the distribution for each choice of $\lambda$ and $\tau$,
and is equal to the Helmholtz free energy $\mathcal{F}_H(\theta') =
\tau \Phi_\tau(\theta')$, where $\Phi_\tau(\theta') = \log \Sigma
\exp(\tau^{-1} \langle \theta', \phi'(x')\rangle)$ is the usual
log-partition function.  Thus, $g(\lambda; \tau)$ is a strictly
convex, analytic function. Using the moment-generating property of
$\Phi_\tau(\theta')$, the gradient of $g(\lambda;\tau)$ is computed
as:
\begin{eqnarray}
\frac{\partial g(\lambda;\tau)}{\partial \lambda_{A,B}} &=& 
    \frac{\partial\Phi_\tau}{\partial\theta'_A} \frac{\partial\theta'_A}{\partial\lambda_{A,B}}
  + \frac{\partial\Phi_\tau}{\partial\theta'_B} \frac{\partial\theta'_B}{\partial\lambda_{A,B}} \nonumber \\
 &=& p_{\lambda,\tau}[\phi_A] - p_{\lambda,\tau}[\phi_B]
\end{eqnarray}
where we use $p[\cdot]$ to denote expectation under $p$. Thus,
appealing to strict convexity, there is a unique $\lambda^*(\tau)$
that minimizes $g(\lambda;\tau)$ and it is also the unique solution of
the set of moment-matching conditions:
\begin{displaymath}
p_{\lambda,\tau}[\phi_A] = p_{\lambda,\tau}[\phi_B], \mbox{ for all } A \equiv B.
\end{displaymath}
These moment-matching conditions are equivalent to requiring that
the marginal distributions $p_{\lambda,\tau}(x_A)$ and
$p_{\lambda,\tau}(x_B)$ are equal for $x_A = x_B$.  We also
note that $\frac{\partial g(\lambda;\tau)}{\partial \tau} =
p_{\lambda,\tau}[-\log p_{\lambda,\tau}]$, which is the \emph{entropy}
of $p_{\lambda,\tau}$ and is positive for all $\lambda$.  Hence, for a
decreasing sequence $\tau_k > 0$ converging to zero, $g(\lambda;\tau)$
converges \emph{monotonically} to $g(\lambda)$.  Likewise, $g^*(\tau_k)$ converges
\emph{monotonically} to $g^*$.

Rather than directly optimizing $g(\lambda)$, we instead perform a
sequence of minimizations with respect to the functions
$g(\lambda;\tau_k)$.  At each step, the previous estimate of
$\lambda_k^* = \arg\min g(\lambda;\tau_k)$ is used to initialize an
iterative method to minimize $g(\lambda;\tau_{k+1})$.  This is
illustrated in Fig. \ref{fig:smoothLR}. At each step, we use the 
following optimization procedure based on the marginal agreement 
condition.

\subsubsection{Iterative Log-Marginal Averaging}

\begin{figure}
\vspace{.3cm}
\hrule
\begin{tabbing}
{\bf ALGORITHM 1 (Discrete LR)}\\ 
It\=erate until convergence:\\
Fo\=r $E \in \calG \mbox{ where } r_E>1$\\
\>Fo\=r $E' \in \calR(E)$\\
\>\>$\hat{f}_{\tau,E'}(x'_{E'}) = \tau \log p_{\tau,\lambda}(x'_{E'})$\\
\>end\\
\>$\bar{f}_{\tau,E}(x_E) = r_E^{-1} \sum_{E'} \hat{f}_{\tau,E'}(x_E)$\\
\>For $E' \in \calR(E)$\\
\>\>$f_{E'}(x_E) \leftarrow f_{E'}(x_E) + \left( \bar{f}_{\tau,E'}(x_E) - \hat{f}_{\tau,E'}(x_E) \right)$\\
\>end\\
end
\end{tabbing}
\vspace{-.2cm}
\hrule
\vspace{-.4cm}
\end{figure}

To minimize $g(\lambda;\tau)$ for a specified $\tau$, starting from an
initial guess for $\lambda$ (or, equivalently, an initial splitting of
$f$), we develop a block coordinate-descent method.  Our approach is
in the same spirit as the iterative proportional fitting procedure
\cite{Ruschendorf95}.

We begin with the case that the augmented model is defined so that no
two replicas of a node are contained in the same connected component
of $\calG'$.  Then, at each step, we minimize over the set of all
Lagrange multipliers associated with features defined within any
replica of $E$.  This is equivalent to solving the condition that the
corresponding marginal distributions $p_{\lambda,\tau}(x'_{E'})$ are
consistent for all $E' \in \calR(E)$.  Algorithm 1 summarizes the
method, which involves computing the log-marginal of each replica
edge, and then updates the functions $f'_{E'}$ according to the rule:
\begin{equation}
f'_{E'}(x_E) \leftarrow f'_{E'}(x_E) + (\bar{f}_{\tau,E}(x_E) - \hat{f}_{\tau,E'}(x_E)) \\
\end{equation}
where
\begin{displaymath}
\hat{f}_{\tau,E'}(x'_{E'}) = \tau \log p_{\lambda,\tau}(x'_{E'}), \;\; \bar{f}_{\tau,E}(x_E) = r_E^{-1} \sum_{E' \in \calR(E)} \hat{f}_{\tau,E'}(x_E).
\end{displaymath}
After the update, the new log-marginals of all replicas $E'$ are equal
to $\bar{f}_{\tau,E}$.  Also, these updates maintain a consistent
representation: $\sum_{E'} (\bar{f}_{\tau,E} - \hat{f}_{\tau,E'}) =
0$.  To handle augmented models with multiple replicas of $E$ in the
same connected subgraph, we only update a \emph{subset} of replicas at
each step, where no two replicas are in the same subgraph. In some
cases, this requires including an extra replica of $E$ to act as an
intermediary in the update step.

Each step of the procedure requires that we compute the marginal
distributions of each replica $E'$ in their respective subgraphs. In
the graphs are thin, these marginals can be computed efficiently, with
computation linear in the size of each subgraph, using standard belief
propagations methods and their junction tree variants.  Moreover, if
we take some care to store the messages computed by belief
propagation, it is possible to amortize the cost of this inference, by
only updating a few ``messages'' at each step.  In fact, it is only
necessary to update those messages along the directed path from the
last updated node or edge to the location in the tree (or junction
tree) of the node or edge currently being updated.  We find that this
generally allows a complete set of updates to be computed with
complexity linear in $n$.  Similar ideas are discussed in
\cite{Kolmogorov05}.

Using Algorithm 1, together with a rule to gradually reduce $\tau$, we
obtain a simple algorithm which generates a sequence $\lambda_k$ such
that $g(\lambda_k)$ converges to $g^*$ and $\lambda_k$ converge to a
point in the set of optimal Lagrange multipliers.

\subsubsection{Iterative Max-Marginal Averaging}

We now consider what happens as $\tau$ approaches zero.  The main
insight is that the (non-normalized) log-marginals converge to
\emph{max-marginals} in the limit as $\tau$ approaches zero:
\begin{equation}
\hat{f}_{\tau,E'}(x'_{E'}) +
g(\lambda,\tau) \rightarrow \hat{f}_{E'}(x'_{E'})
\triangleq \max_{x'_{\setminus E'}} f'(x'_{E'},x'_{\setminus E'};\lambda)
\end{equation}
Hence, as $\tau$ becomes small, the marginal agreement conditions are
similar to a set of \emph{max-marginal agreement} conditions among all
replicas of an edge or node.  One could consider a ``zero-temperature''
version of Algorithm 1 aimed at solving these max-marginal
conditions directly:
\begin{eqnarray}\label{eq:max_marg_match}
f'_{E'}(x_E) &\leftarrow& f'_{E'}(x_E) + \left(\bar{f}_E(x_E) -
\hat{f}_{E'}(x_E) \right) \nonumber\\ \bar{f}_E(x_E) &=& r_E^{-1}
\sum_{E'} \hat{f}_{E'}(x_E)
\end{eqnarray}
Here, $\bar{f}_E$ is the averaged max-marginal over all replicas of
$E$.  Note that $\hat{f}_{E'}(x_E) \ge \hat{f}_E(x_E) \triangleq
\max_{x_{\setminus E}} f(x)$ for all $x_E$ and $E' \in
\mathcal{R}(E)$, which implies $\bar{f}_E(x_E) \ge \hat{f}_E(x_E)$.
This ``zero-temperature'' approach has close ties to max-sum diffusion
(see \cite{Werner07} and reference therein) and Kolmogorov's serial
approach to TRMP \cite{Kolmogorov05}.

In our framework, one can show that $\lambda^* \triangleq \lim_{\tau
\rightarrow 0} \lambda^*(\tau)$ is well-defined and minimizes
$g(\lambda)$.  This point $\lambda^*$ also satisfies the max-marginal
agreement condition and is therefore a fixed point of max-marginal
averaging.  However, the max-marginal agreement condition by itself
does not uniquely determine $\lambda^*$ and, in fact, is not
sufficient to insure that $g(\lambda)$ is minimized (this is related
to the existence of non-minimal fixed-points observed by Kolmogorov).
Hence, our approach to minimize $g(\lambda;\tau)$ while gradually
reducing the temperature has the advantage that it cannot get stuck in
such spurious fixed-points.  It also helps to accelerate convergence,
because the initial optimization at higher temperatures serves to
smooth over irregularities of the dual function.

\begin{figure*}
\centering
\epsfig{file=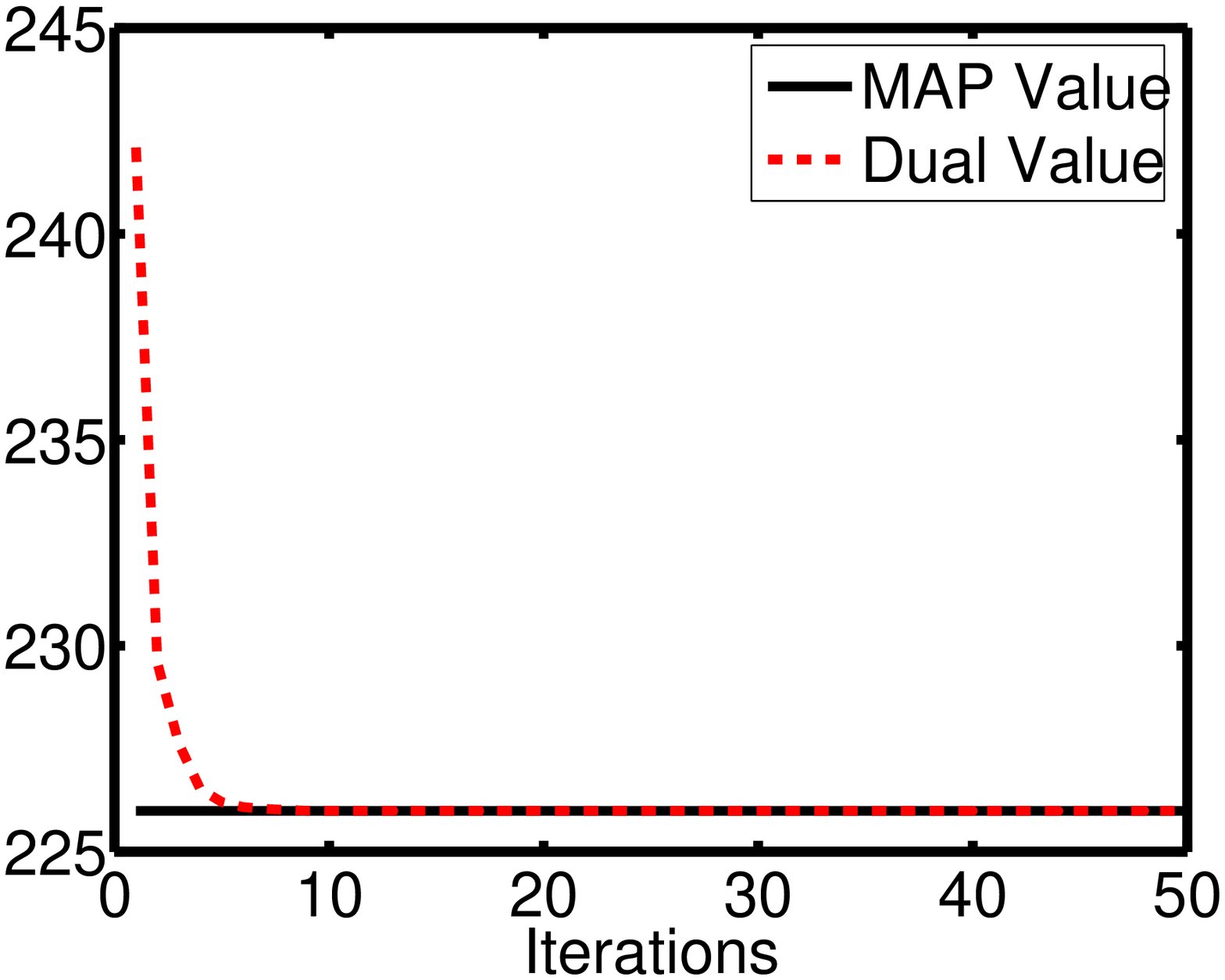,scale=0.17}
\epsfig{file=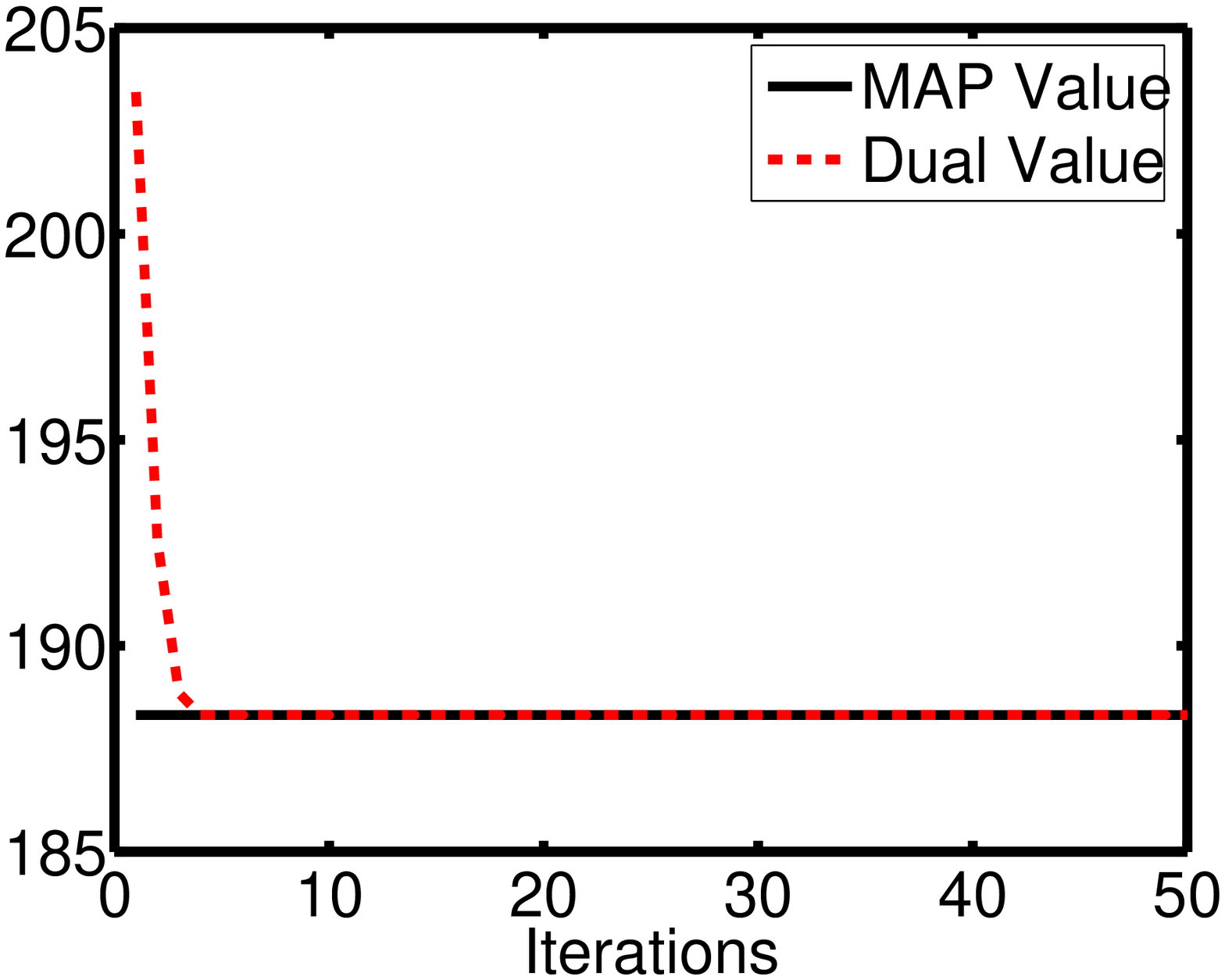,scale=0.17}
\epsfig{file=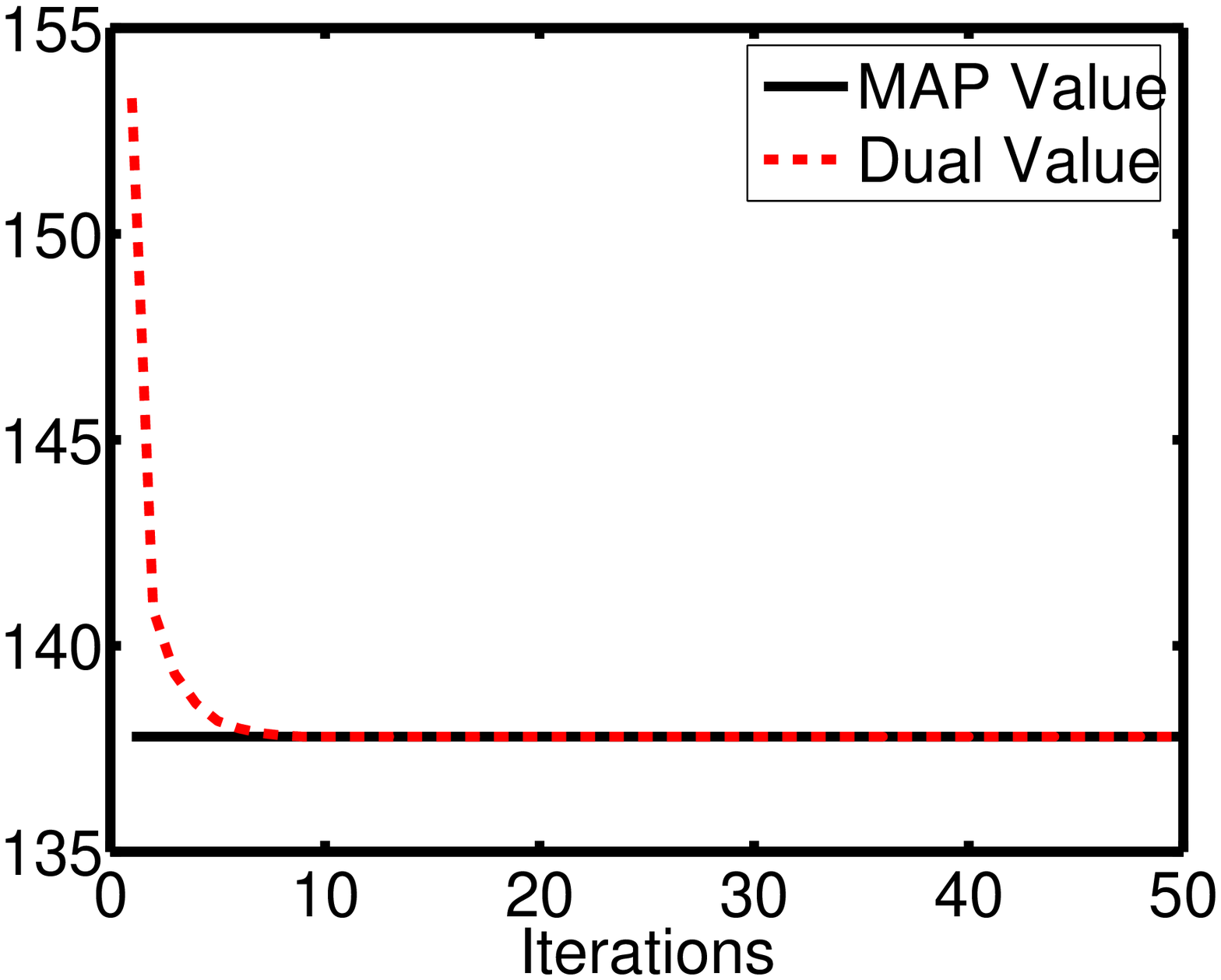,scale=0.17}
\epsfig{file=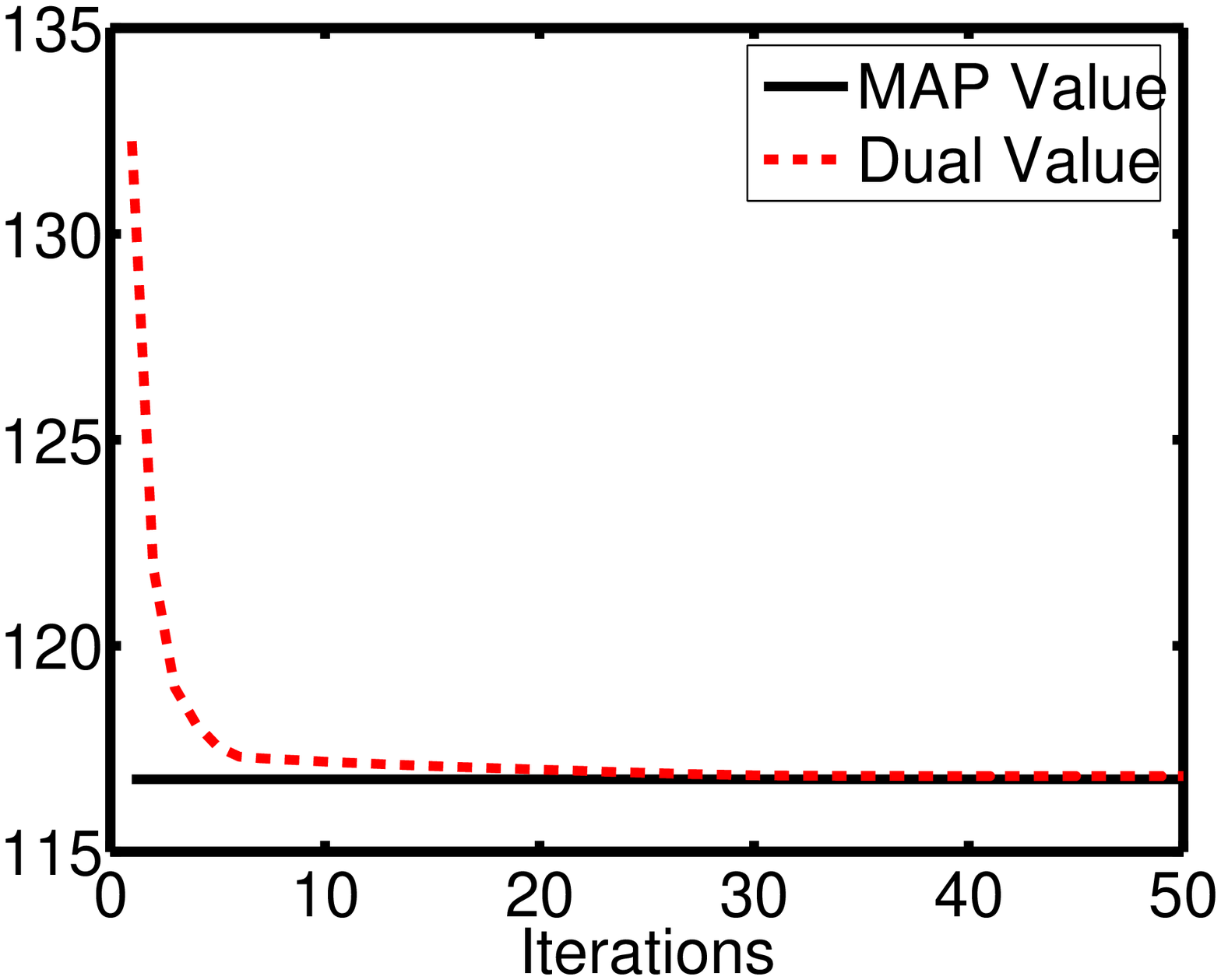,scale=0.17}
\epsfig{file=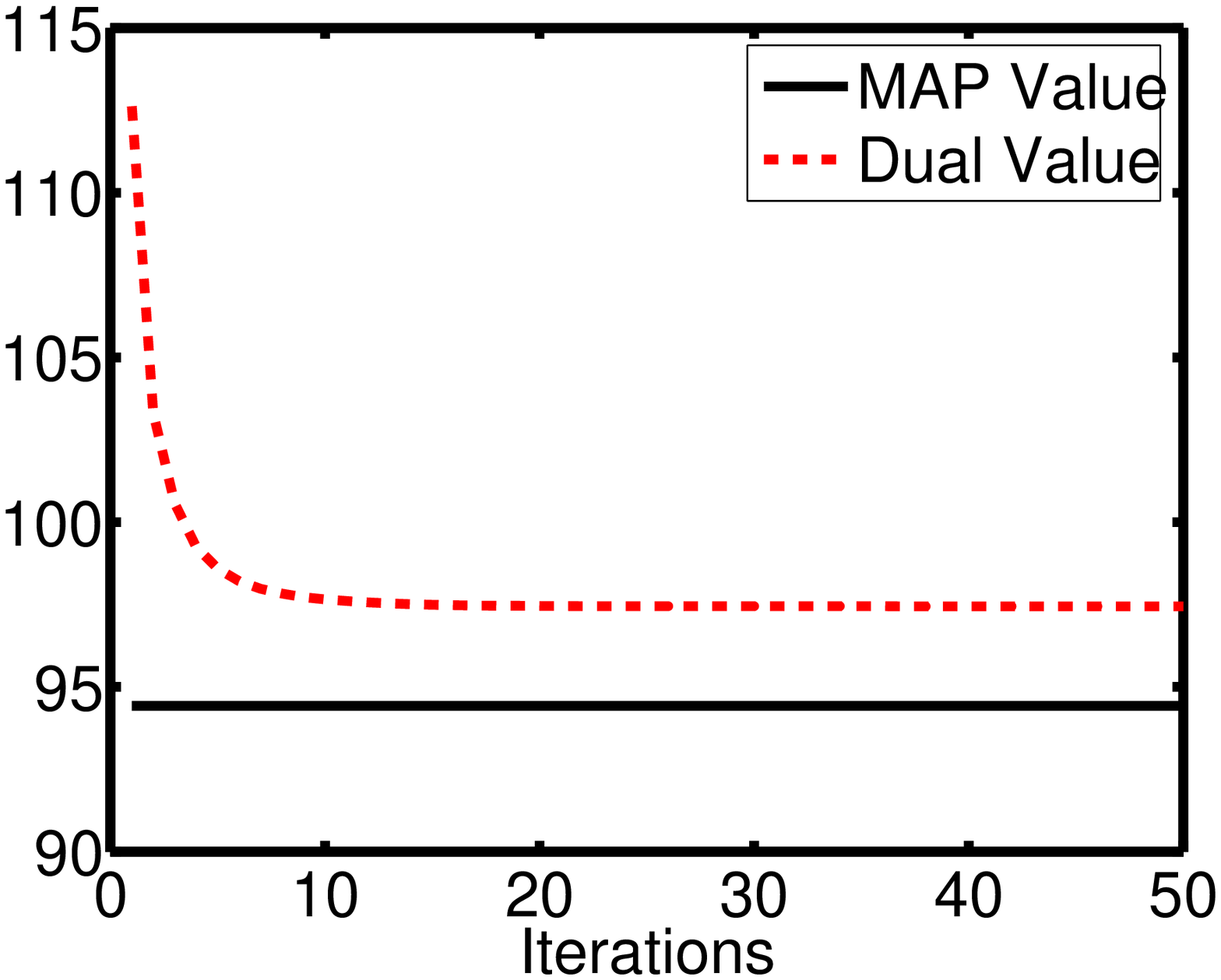,scale=0.17}\\
\comment{\epsfig{file=attractive3_max_marg_err.eps,scale=0.17}
\epsfig{file=attractive1_max_marg_err.eps,scale=0.17}
\epsfig{file=frustrated1_max_marg_err.eps,scale=0.17}
\epsfig{file=frustrated4_max_marg_err.eps,scale=0.17}
\epsfig{file=frustrated2_max_marg_err.eps,scale=0.17}\\}
\epsfig{file=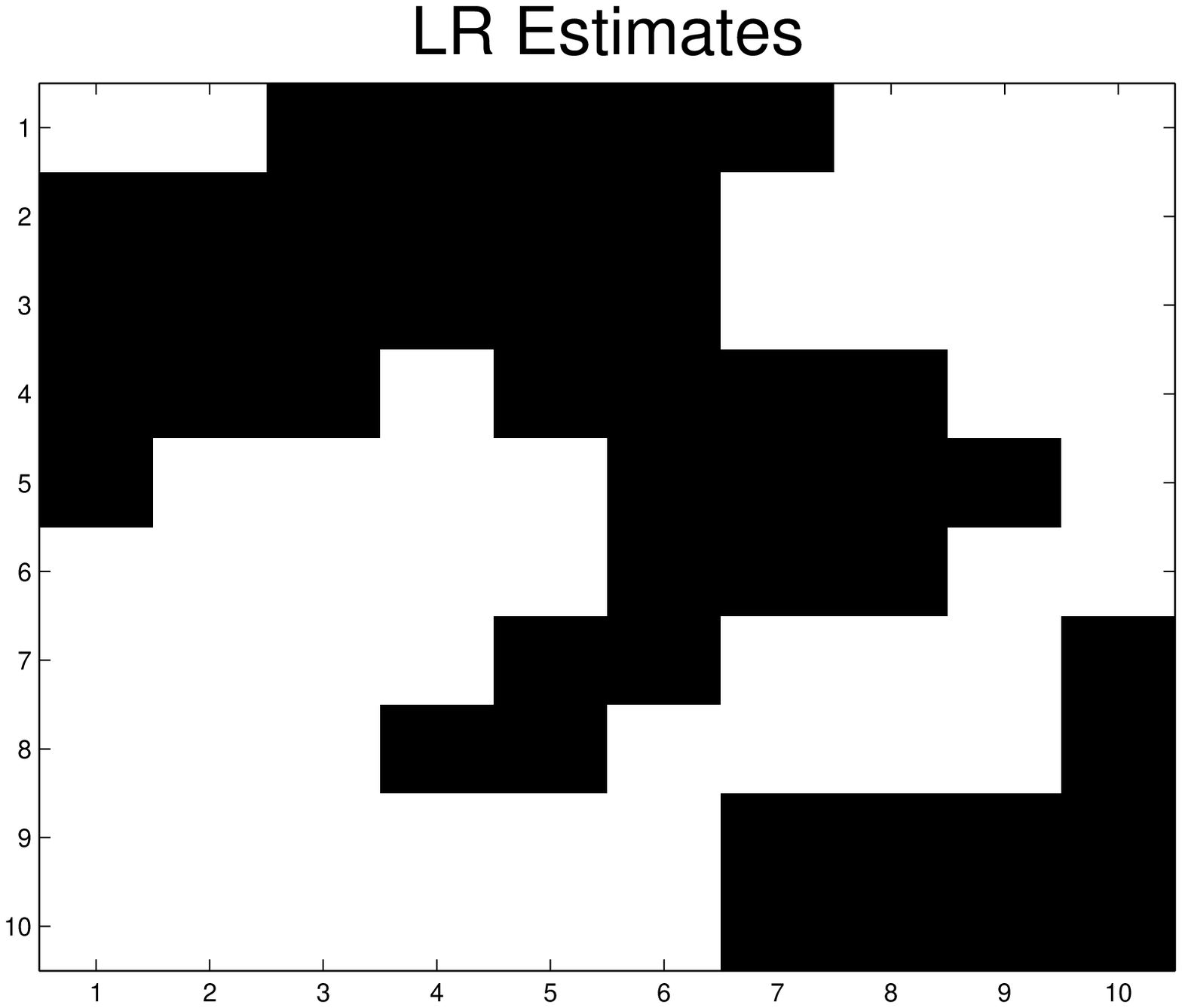,scale=0.18}\hspace{.1cm}
\epsfig{file=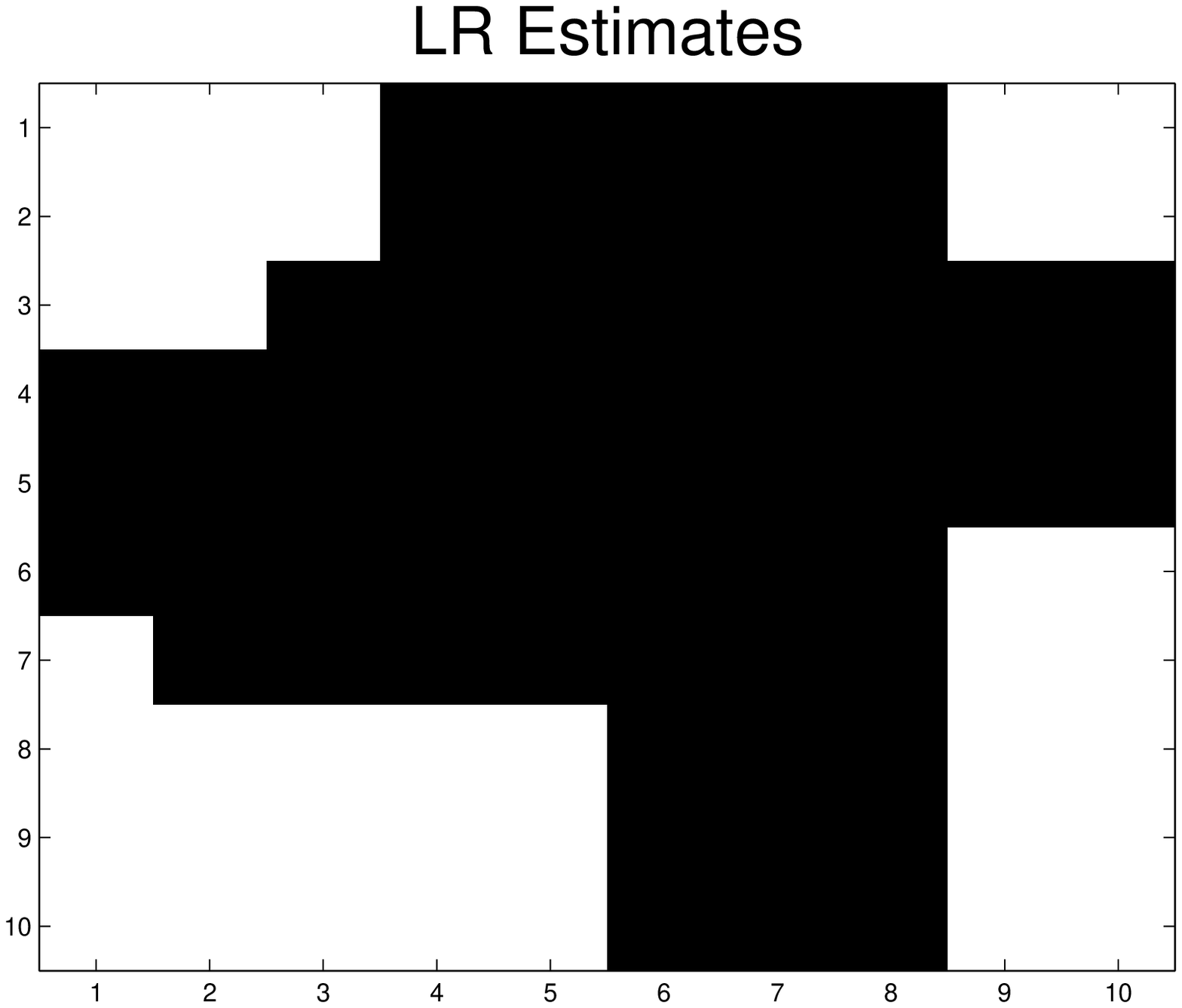,scale=0.18}\hspace{.1cm}
\epsfig{file=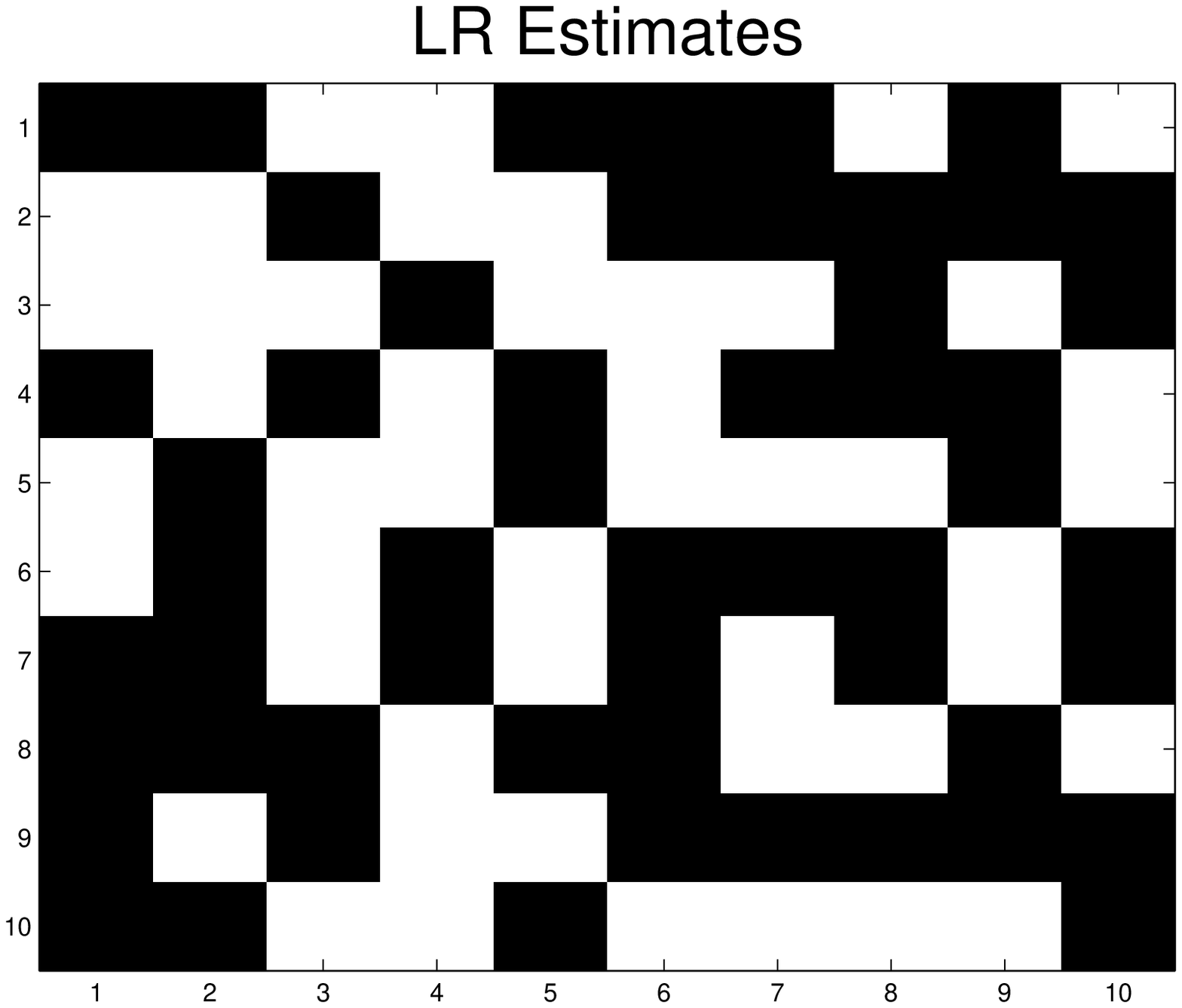,scale=0.18}\hspace{.1cm}
\epsfig{file=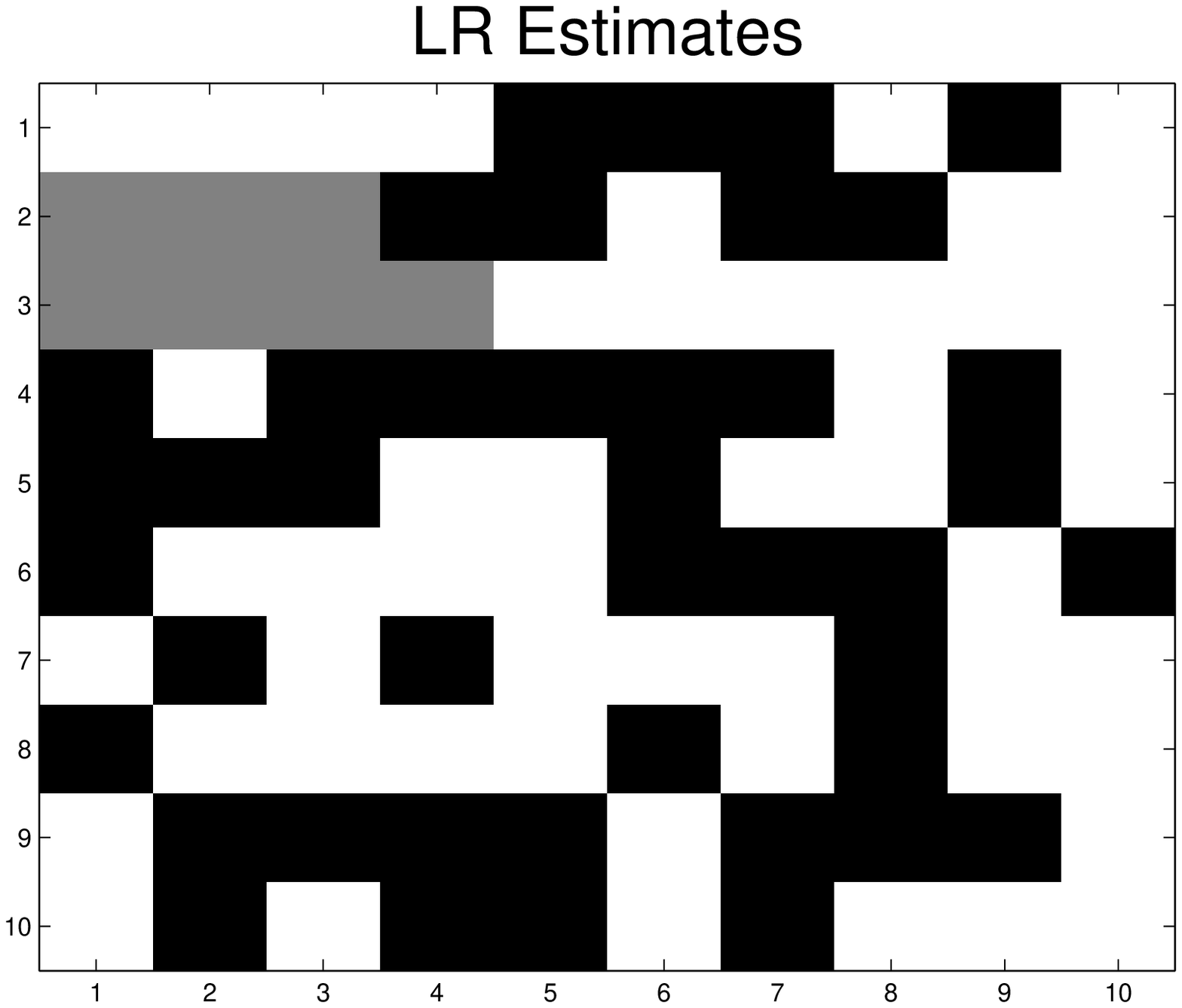,scale=0.18}\hspace{.1cm}
\epsfig{file=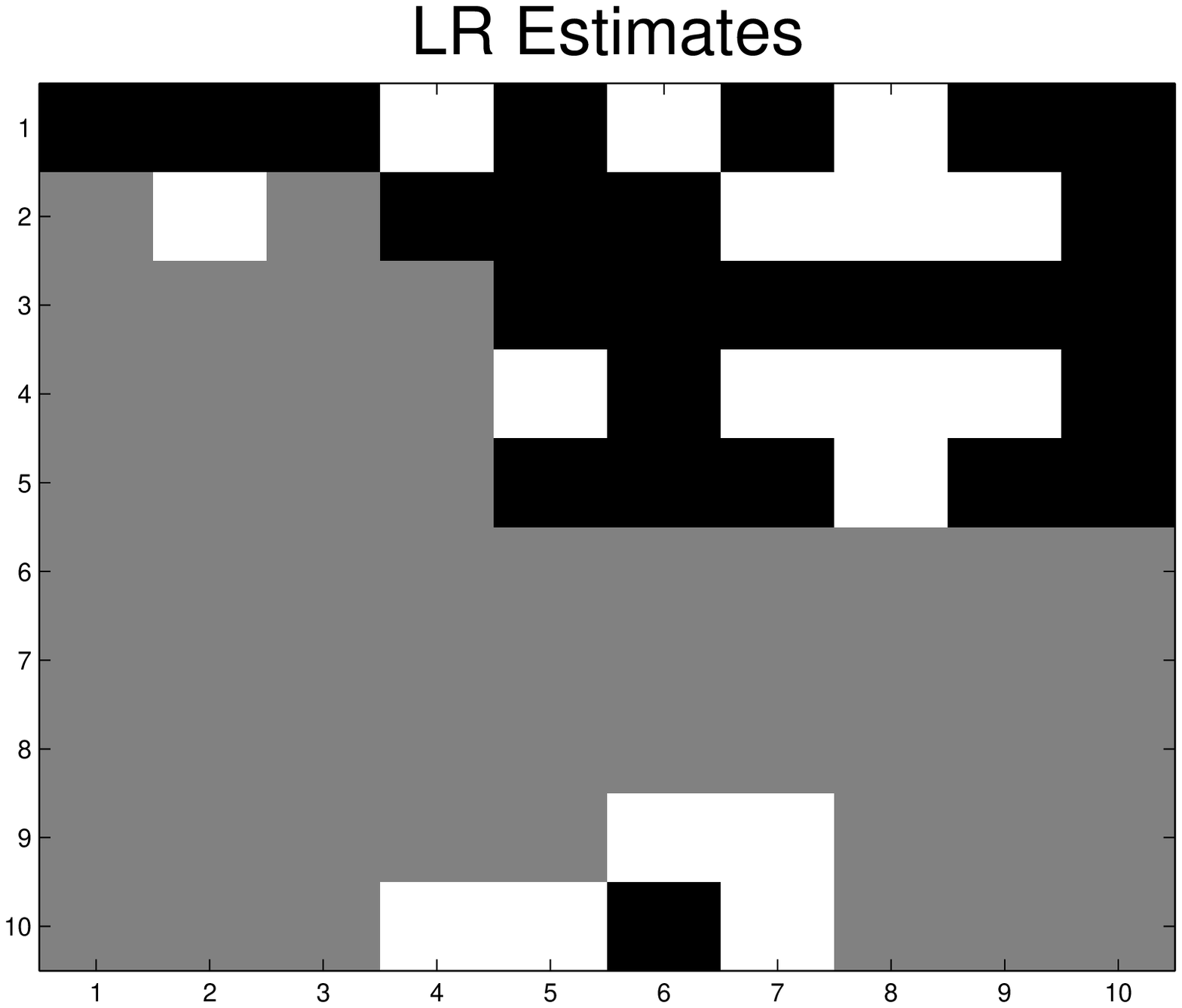,scale=0.18}
\caption{\label{fig:discreteLR} Five examples for discrete LR showing:
(top row) convergence of $g(\lambda)$ to $g^*$ compared to $f^*$
(horizontal line); (bottom row) the resulting estimates generated by
relaxed max-marginals (grey areas denote non-unique maximum).  The
first two columns are examples of attractive models with $\sigma=2
\mbox{ and } 1$.  The last three columns are frustrated models with
$\sigma= 1.5, 1, \mbox{ and } .7$.}
\vspace{-.4cm}
\end{figure*}

\emph{Computational Examples.} In this section we provide some
preliminary results using our approach to solve binary MRFs. These
examples are for a binary model $x_v \in \{-1,+1\}$ defined on a $10
\times 10$ grid similar to the one seen in Fig. \ref{fig:graphs}(a).
For each node, we include a node potential $f_v(x_v) = \theta_v x_v$
with $\theta_v \sim N(0,\sigma^2)$.  For each edge, we include an edge
potential $f_{u,v}(x_u,x_v) = \theta_{uv} x_u x_v$ with $\theta_{uv} =
1$ in the ``attractive'' model and random $\theta_{uv} = \pm 1$ in the
``frustrated'' model.  Hence, $\sigma$ controls the strength of node
potentials relative to edge potentials.  As seen in
Fig. \ref{fig:discreteLR}, we obtain strong duality $g^*=f^*$ and
recover the correct MAP estimates in attractive models. This is
consistent with a result on optimality of TRMP in attractive models
\cite{KolmogorovWainwright05}. In the frustrated model, the same holds
with strong node potentials, but as $\sigma$ is decreased the
frustration of the edge potentials cause a duality gap. However, even
in these cases, we have observed that some nodes have a unique maximum
in their re-summed max-marginals, and these nodes provide a partial MAP
estimate that agrees with the correct global MAP estimate.  This is
apparently related to the \emph{weak tree agreement} condition for
partial optimality in TRMP \cite{KolmogorovWainwright05}. 

\section{Gaussian Lagrangian Relaxation}

In this section we apply the LR approach to the
problem of MAP estimation in Gaussian graphical models, which
is equivalent to maximizing a quadratic objective function
\begin{equation}
f(x;h,J) = -\frac{1}{2} x^T J x + h^T x,
\end{equation}
where $J \succ 0$ is sparse with respect to $\calG$.  Again, we
construct an augmented model, which is now specified by an information
form $(h',J')$, defined by a larger graph $\calG'$.  For consistency,
we also require $f'(\zeta(x);h',J')=f(x;h,J)$ for all $x$.  Denoting
variable replication by $\zeta(x) = A x$, this is equivalent to $A^T
J' A = J$ and $A^T h' = h$. In order for the dual function to be
well-defined, we also require that $J' \succ 0$.  For general $J \succ
0$, it is possible that, for a given augmented graph $\calG'$, there
do not exist any $J' \succ 0$ defined on $\calG'$ such that $A^T J' A
= J$.  To avoid this issue, we will focus on models that are of the
form:
\begin{equation}\label{eq:e}
f(x) = \sum_{E \in \calF} f_E(x_E)
\end{equation}
where $\calF$ is a hyper-graph, composed of cliques of $\calG$, and
each term $f_E(x_E)$ is itself a quadratic form $f_E(x_E) =
-\frac{1}{2} x_E^T J_E x_E + h_E^T x_E$ based on $J_E \succ 0$.  Then,
$J = \sum_E [J_E]_V$ is the sum of these (zero-padded)
submatrices. Then, it is simple to obtain a valid augmented model.  We
split each $J_E$ between its replicas as $J_{E'} = r_E^{-1} J_E$ to
obtain $J' = \sum_{E' \in \calF'} [J_{E'}]_{V'} \succ 0$.

If there exists a representation of $J$ in terms of $2 \times 2$
\emph{pairwise} interactions $J_E \succ 0$, it is said to be
\emph{pairwise normalizable}.  This condition is equivalent to the
walk-summability condition considered in \cite{Malioutov*06}, which is
related to the convergence (and correctness) of a variety of
approximate inference methods \cite{Malioutov*06,Chandrasekaran*07}.
Here, we show that for the more general class of models of the form
(\ref{eq:e}), we obtain a convergent iterative method for solving the
dual problem that is tractable provided the cliques are not too large.
Moreover, for this class of Gaussian models, we show that there is
\emph{no duality gap} and we always converge to the unique MAP
estimate of the model.  As an additional bonus, we also find that, by
solving marginal agreement conditions in the augmented Gaussian model,
we obtain a set of upper-bounds on the variances of each variable,
although these bounds are often rather loose.

\subsection{Gaussian LR with Linear Constraints}

We begin by considering the Lagrangian dual of the 
following linearly-constrained quadratic program:
\begin{equation}
\begin{array}{ll}
\mbox{maximize} & -\tfrac{1}{2} x'^T J' x' + h'^T x'\\
\mbox{subject to} & x'_a = x'_b \mbox{ for all } a \equiv b.
\end{array}
\end{equation}
We may express the linear constraints on $x'$ as $H x' =
0$. Relaxing these constraints leads to the following dual function:
\begin{eqnarray}
g(\lambda) &=& \max_{x'} \{ -\tfrac{1}{2} x'^TJ'x'+(h'+H^T\lambda)x' \} \nonumber \\
           &=& \tfrac{1}{2} (h'+H^T \lambda)^T J'^{-1} (h'+H^T \lambda)
\end{eqnarray}
Moreover, by strong duality of quadratic programming \cite{Bertsekas95}, it holds
that $g^*=f^*$. We also note the following equivalent representation
of the dual problem:
\begin{equation}\label{eq:qp1}
g^* = \left\{
\begin{array}{ll}
\mbox{minimize} & \tfrac{1}{2} h'^T J'^{-1} h'\\
\mbox{subject to} & A^T h' = h \\
\end{array}
\right.
\end{equation}
Here, $h'$ is the problem variable, and we consider all possible
choices of $h'$ that are consistent with $h$ under the constraint $x'
= A x$.  The optimal choice of $h'$ in this problem is the one which
leads to consistency in the estimate $\hat{x}' = J'^{-1} h'$.

\subsection{Quadratic Constraints and Log-Det Regularization}

Although, in Gaussian models, it is sufficient to include only linear
constraints (there is no duality gap), our method can also accommodate
quadratic constraints, and this results in faster convergence and
tighter bounds on variances. Consider the constrained optimization
problem:
\begin{equation}
\begin{array}{ll}
\mbox{maximize} & -\tfrac{1}{2} x'^T J' x' + h'^T x'\\
\mbox{subject to} 
& x_a = x_b, x_a^2 = x_b^2 \mbox{ for all } a \equiv b,\\
& x_{a_1}x_{a_2} = x_{b_1}x_{b_2} \mbox{ for all } (a_1,a_2) \equiv (b_1,b_2).
\end{array}
\end{equation}
This leads to the following equivalent version of the dual
problem with problem variables $(h',J')$:
\begin{equation}
\begin{array}{ll}
\label{eq:c}
\mbox{minimize} & \tfrac{1}{2} h'^T J'^{-1} h' \\
\mbox{subject to} & A^T h'=h, \; A^T J' A = J, \; J' \succ 0.
\end{array}
\end{equation}
Any solution of the linearly-constrained relaxation provides a
feasible point for this problem, so the value of (\ref{eq:c}) is
less than or equal to that of (\ref{eq:qp1}).  However, since there is
no duality gap in (\ref{eq:qp1}), the value of the two problem
are equal, both achieve $g^*=f^*$ and obtain the MAP estimate.

While the choice of $J'$ does not affect the value of the dual
problem, it does effect variance estimates and convergence of
iterative methods.  Hence, we regularize the choice of $J'$ by adding
a penalty $-\tfrac{1}{2} \log\det J'$ to the objective of
(\ref{eq:c}), which also serves as a barrier function enforcing $J'
\succ 0$.  The resulting objective function is then equivalent to
$\Phi(h,J)$, which shows a parallel to our earlier approach for
``smoothing'' the dual function in discrete
problems. \comment{(although, here, there is no need for a temperature
parameter). Similarly, we find that minimizing the log-partition
function in the Gaussian model, subject to consistency constraints,
reduces to matching means and covariances among replicas of a node or
edge.}

\subsection{Gaussian Moment-Matching}

\begin{figure}
\vspace{.3cm}
\hrule
\begin{tabbing}
{\bf ALGORITHM 2 (Gaussian LR)}\\
It\=erate until convergence:\\
Fo\=r $E \in \calG \mbox{ where } r_E>1$\\
\>Fo\=r $E' \in \calR(E)$\\
\>\> Compute moments $(\hat{x}_{E'},P_{E'})$ in $(h',J')$.\\
\>\> $\hat{J}_{E'} = P_{E'}^{-1}, \; \hat{h}_{E'} = P_{E'}^{-1} h_{E'}$\\
\>end\\
\>$\bar{J}_E = r_E^{-1} \sum_{E'} \hat{J}_{E'}, \; \bar{h}_E = r_E^{-1} \sum_{E'} \hat{h}_{E'}$\\
\>For $E' \in \calR(E)$\\
\>\>$J'_{E',E'} \leftarrow J'_{E',E'} + \left(\bar{J}_E - \hat{J}_{E'} \right)$\\
\>\>$h'_{E'} \leftarrow h'_{E'} + \left(\bar{h}_E - \hat{h}_{E'} \right)$\\
\>end\\
end
\end{tabbing}
\vspace{-.2cm}
\hrule
\vspace{-.4cm}
\end{figure}

We develop an approach in the same spirit as the Gaussian iterative
scaling method \cite{SpeedKiiveri86}. We minimize the log-partition
function with respect to the information parameters over all replicas
of a node or edge, subject to consistency and positive definite
constraints.  The optimality condition for this minimization is that
the marginal moments (means and variances) of all replicas are
equalized.  It can be shown that the following information-form
updates achieve this objective.  First, for all replicas $E'$ of $E$,
we compute the marginal information parameters given by sparse
Gaussian elimination of $C = V' \setminus E'$ in $(J',h')$:
\begin{eqnarray}
\hat{J}_{E'} &=& J'_{E',E'} - J'_{E',C} (J'_{C,C})^{-1} J'_{C,E'} \nonumber \\
\hat{h}_{E'} &=& h'_{E'} - J'_{E',C} (J'_{C,C})^{-1} h'_{C}
\end{eqnarray}  
This is equivalent to $\hat{J}_{E'} = P_{E'}^{-1}$ and $\hat{h}_{E'} =
P_{E'}^{-1} \hat{x}_{E'}$. Next, we average these marginal information
forms over all replicas:
\begin{equation}\label{eq:marg_info_match}
\bar{J}_E = r_E^{-1} \sum_{E'} \hat{J}_{E'}, \;\; \bar{h}_E = r_E^{-1} \sum_{E'} \hat{h}_{E'}
\end{equation}
Finally, we update the information form according to:
\begin{eqnarray}
J'_{E',E'} &\leftarrow& J'_{E',E'} + (\bar{J}_E - \hat{J}_{E'}) \nonumber \\
h'_{E'} &\leftarrow& h'_{E'} + (\bar{h}_E - \hat{h}_{E'})
\end{eqnarray}
Using the characterization of positive-definiteness of a block matrix
in terms of a principle submatrix and its Schur complement, it can be
shown that this update preserves positive definiteness of $J'$.  It
also preserves consistency, e.g., $\sum_{E'} (\bar{J}_E -
\hat{J}_{E'}) = 0$.  After the update, the new marginal information
parameters for all replicas of $E$ are equal to
$(\bar{h}_E,\bar{J}_E)$.  Algorithm 2 summarizes this iterative
approach for solving the Gaussian LR problem.

Lastly, using the fact that $\bar{f}_E(x_E) \ge \hat{f}_E(x_E)$ for
all $x_E$ and that there is no duality gap upon convergence, we
conclude that the final equalized marginal information must satisfy
$\bar{J}_E \preceq \hat{J}_E \triangleq J_{E,E} - J_{E,\setminus
E}(J_{\setminus E,\setminus E})^{-1}J_{\setminus E,E}$. Hence, LR
gives an upper-bound on the true variance: $P_E=(\hat{J}_E)^{-1}
\preceq (\bar{J}_E)^{-1}$.  If each replica of $E$ is contained in a
separate connected component of $\calG'$, then a tighter bound 
holds: $P_E \preceq (r_E \bar{J}_E)^{-1}$.

\begin{figure}
\centering
\epsfig{figure=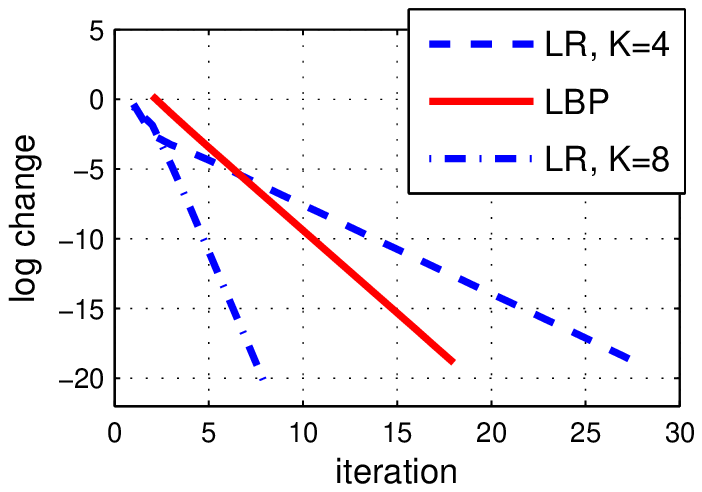,scale=.55}
\epsfig{figure=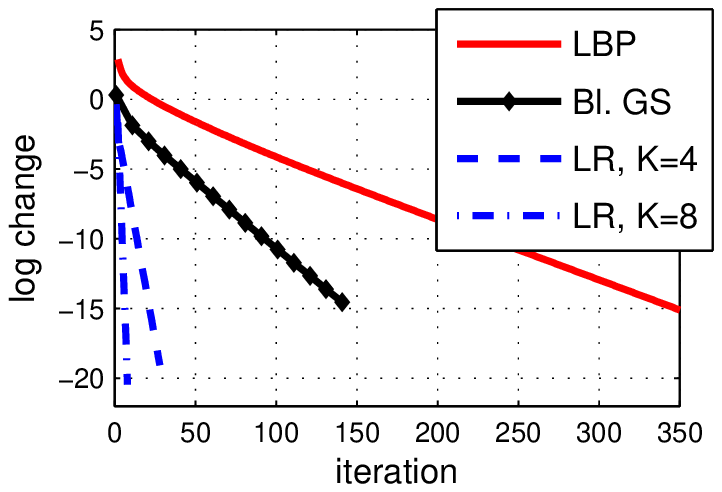,scale=.55}\\
\epsfig{figure=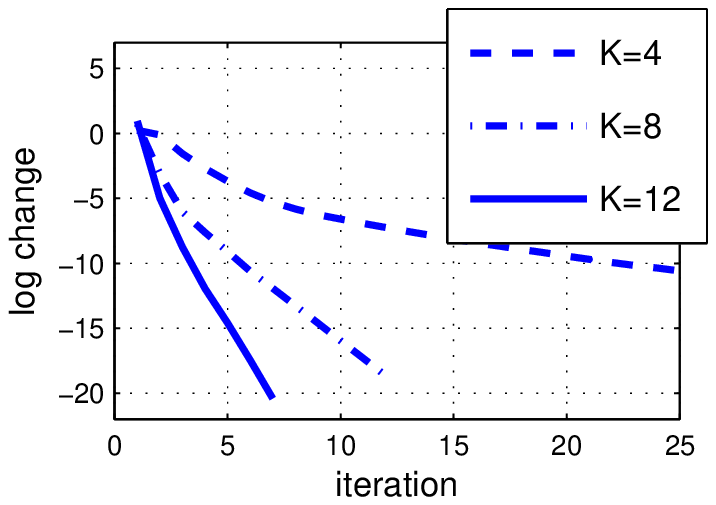,scale=.55}
\epsfig{figure=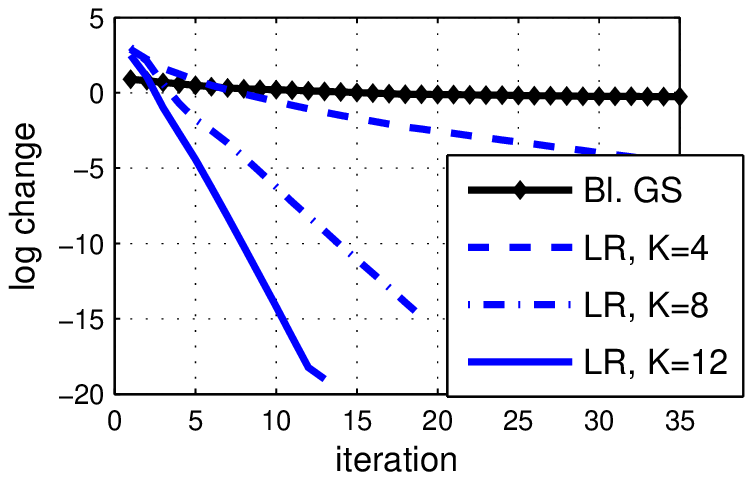,scale=.55}
\caption{\label{fig:gauss_LR} Convergence plots for variances (left) and
means (right), in the thin-membrane model (top) and thin-plate model (bottom).}
\vspace{-.3cm}
\end{figure}

\emph{Computational Examples.} We apply LR for two Gaussian models
defined on a $50 \times 50$ 2D grid with correlation lengths
comparable to the size of the field. First, we use the \emph{thin
membrane} model, which encourages neighboring nodes to be similar by
having potentials $f_{ij} = (x_i - x_j)^2$ for each edge
$\{i,j\} \in \calG$. We split the 2D model into vertical strips of
narrow width $K$, which have overlap $L$ (we vary $K$ and set
$L=2$). We impose marginal agreement conditions in $K \times L$ blocks
in these overlaps. The updates are done consecutively, from top to
bottom blocks, from the left to the right strip. A full update of all
the blocks constitutes one iteration. We compare LR to loopy belief
propagation (LBP). The LBP variances are underestimates by $21.5$
percent (averaged over all nodes), while LR variances for $K=8$ are
overestimates by $16.1$ percent. In Figure \ref{fig:gauss_LR} (top) we
show convergence of LR for several values of $K$, and compare it to
LBP. The convergence of variances is similar to LBP, while for the
means LR converges considerably faster. In addition, the means in LR
converge faster than using block Gauss-Seidel on the same set of
overlapping $K \times 50$ vertical strips.

Next, we use the \emph{thin plate model}, which enforces that each
node $v$ is close to the average of its nearest neighbors $N(v)$ in
the grid, and penalizes curvature. At each node there is a potential:
$f_i(x_i,x_{N(i)}) = (x_i - \tfrac{1}{|N(i)|} \sum_{j \in N(i)}
x_j)^2$. LBP does not converge for this model. LR gives rather loose
variance bounds for this more difficult model: for $K=12$, it
overestimates the variances by $75.4$ percent. More importantly, it
accelerate convergence of the means. In Figure \ref{fig:gauss_LR}
(bottom) we show convergence plots for means and variances, for
several values of $K$. As $K$ increases, the agreement is achieved
faster, and for $K=12$ agreement is achieved in under $13$ iterations
for both means and variances. We note that LR with $K=4$ converges
much faster for the means than block Gauss-Seidel.

\section{Multi-Scale Lagrangian Relaxation}

In this section, we propose an extension of the LR method considered
thus far.  Previously, we have considered relaxations based on
augmented models where $x' = \zeta(x)$ involves replication of
variables.  Here, we consider more general definition of $\zeta$ to
allow the augmented model to include \emph{summary variables}, such as
a sum over a subset of variables, or any linear combination of these.
In discrete models, summary variables can also be non-linear functions
of $x$. For example, ``parity bits'' are used in coding
applications and the ``majority rule'' is used to define
coarse-scale binary variables in the renormalization group approach
\cite{Gidas89}.

Using this idea, we develop a \emph{multiscale} Lagrangian relaxation
approach for MRFs defined on grids.  The purpose of this relaxation is
similar to that of the multigrid and renormalization group methods
\cite{Trottenberg*01,Gidas89}. Iterative methods generally involve
simple rules that propagate information locally within the graph.  Using a
multiscale representation of the model allows information to
propagate through coarse scales, which improves the rate of
convergence to global equilibrium.  Also, in discrete problems,
such multiscale representations can help to avoid local minima.  In
the context of our convex LR approach, we expect this to translate
into a reduction of the duality gap to obtain the optimal MAP estimate
in a larger class of problems.

\begin{figure}[t]
\centering
(a)\epsfig{file=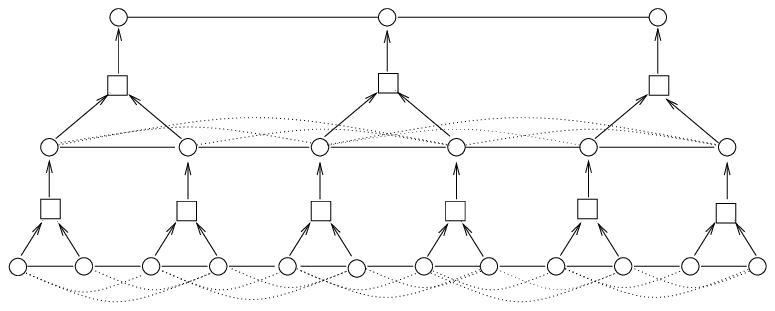,scale=0.8}\\
\vspace{.2cm}
(b)\epsfig{file=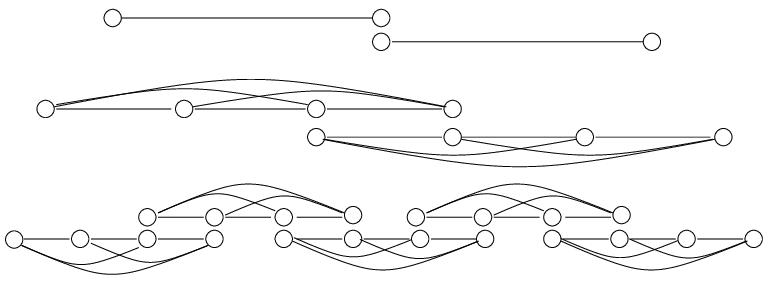,scale=0.8}
\caption{\label{fig:multiscaleLR} Illustration of multiscale
LR method. (a) First, we define an equivalent
multiscale model subject to cross-scale constraints. Relaxing these
constraints leads to a set of single-scale models. (b) Next, each single
scale is relaxed to a set of tractable subgraphs.}
\vspace{-.4cm}
\end{figure}

\subsection{An Equivalent Multiscale Model}

We illustrate the general idea with a simple example based on a 1D
Markov chain.  While this case is actually tractable by exact methods,
it serves to illustrate our approach, which generalizes to
2D grids and 3D lattices.  In Fig. \ref{fig:multiscaleLR}, we show how
to construct the augmented model $f'(x')$ defined on a graph $\calG'$.
This is done in two stages.

First, as illustrated in Fig. \ref{fig:multiscaleLR}(a), we introduce
coarse-scale representations of the fine scale variables by
recursively defining summary variables at coarser scales to be
functions of variables at the next level down.  This defines a set of
cross-scale constraints, denoted by the square nodes. To allow
interactions between coarse-scale variables, while maintaining
consistency with the original single-scale model, we introduce extra
edges (the dotted ones in Fig. \ref{fig:multiscaleLR}(a)) between
blocks of nodes that have a (solid) edge between their summary nodes
at the next coarser scale.  This representation allows us to define a
family of constrained multiscale models that are all equivalent to the
original single-scale model.  For 2D and 3D lattices, this model is
still intractable even after relaxing the cross-scale constraints
because each scale is itself intractable.

Next, to obtain a tractable dual problem, we break up the graph into smaller
subgraphs, introducing additional constraints to enforce consistency
among replicated variables.  In the example, we break the augmented
graph at each scale into its maximal cliques, shown in
Fig. \ref{fig:multiscaleLR}(b).  This defines the final augmented
model and the corresponding graph. In a 2D graph, the same idea
applies, but we obtain a set of maximal cliques consisting of
overlapping $2 \times 4$ and $4 \times 2$ blocks of the grid.
Alternatively, we could break up the 2D grid into a set of width 2
vertical strips, as discussed previously.

Now, the procedure is essentially the same as before. We start with
the equivalent constrained optimization problem defined on the
augmented graph, now subject to both in-scale and cross-scale
constraints. We obtain a tractable problem by introducing Lagrange
multipliers to relax these constraints.  Then we iteratively adjust
the Lagrange multipliers to minimize the dual function, with the aim
of eliminating constraint violations to obtain the desired MAP
estimate.  This is equivalent to adjusting the augmented model
$f'(x')$ on $\calG'$, subject to the constraint that it remains
equivalent to $f(x)$ for all $x' = Ax$.

\subsection{Gaussian Multiscale Moment-Matching}

We demonstrate this approach in the Gaussian model.  To carry out the
minimization, we again use a block coordinate-descent method that
finds an exact minimum over a subset of Lagrange multipliers at each
step. The replica constraints are handled the same as before. Here, we
briefly summarize our approach to handle the cross-scale summary
constraints. Let $x_1$ and $x_2$ denote two random vectors at
consecutive scales coupled by the constraint $x_2 = A x_1$.  Let
$(\hat{h}_1,\hat{J}_1)$ and $(\hat{h}_2,\hat{J}_2)$ denote their corresponding \emph{marginal}
information parameters. Relaxing the constraints $x_2 = A x_1$ and
$x_2 x_2^T = A x_1 x_1^T A^T$, with Lagrange multipliers
$(\lambda,-\tfrac{1}{2}\Lambda)$, leads to the following optimality
conditions:
\begin{eqnarray}
(\hat{J}_2+\Lambda)^{-1} &=& A (\hat{J}_1-A^T\!\Lambda\, A)^{-1} A^T \\
(\hat{J}_2+\Lambda)^{-1}(\hat{h}_2+\lambda) &=& A (\hat{J}_1-A^T\!\Lambda\, A)^{-1}(\hat{h}_1-A^T\lambda) \nonumber
\end{eqnarray}
We find that the solution is:\footnote{The formula (\ref{eq:multiscale_update}) corresponds to a generalization of Algorithm 2, in which the moments $(\hat{x}_1,P_1)$ of fine-scale variables $x_1$ are replaced by the corresponding moments $(A \hat{x}_1, A \hat{P}_1 A^T)$ of the summary statistic $\tilde{x}_1 = A x_1$.}
\begin{eqnarray}\label{eq:multiscale_update}
\Lambda &=& \tfrac{1}{2} \{(A\hat{J}_1^{-1}A^T)^{-1} - \hat{J}_2\} \nonumber\\
\lambda &=& \tfrac{1}{2} \{(A\hat{J}_1^{-1}A^T)^{-1} A \hat{J}_1^{-1} \hat{h}_1 - \hat{h}_2\}
\end{eqnarray}
The model $(h',J')$ is then updated by adding $(\lambda,\Lambda)$ to
the coarse-scale and subtracting $(A^T \lambda,A^T \!\Lambda\, A)$ from the
fine scale. This update enforces the moment conditions $\hat{x}_2 = A
\hat{x}_1$ and $P_2 = A P_1 A^T$ while maintaining consistency of the
model $(h',J')$. Similar updates can be derived when there are
multiple replicas of $x_1$ and $x_2$.  These methods, together with
those described previously, are used to minimize the dual function in
the Gaussian multiscale relaxation.

\begin{figure}
\centering
\epsfig{file=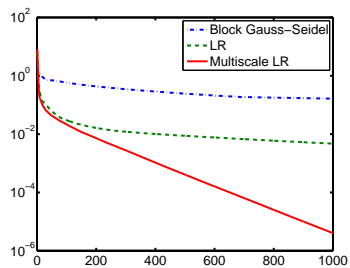,scale=.25}
\caption{\label{fig:multiscale_lr_example}Convergence of single- and multi-scale LR and block Gauss-Seidel.}
\vspace{-.4cm}
\end{figure}

\emph{Multiscale Example.} We provide a preliminary result involving a
1D thin-membrane model with $1024$ nodes.  It is defined to have a
long correlation length comparable to the length of the field. Using a
random $h$-vector, we solve for the MAP estimates using three methods:
a standard block Gauss-Seidel iteration using overlapping blocks of
size 4; the (single-scale) Gaussian LR method with the same choice of
blocks; and the multiscale LR method.  The convergence of all three
methods are shown in Fig. \ref{fig:multiscale_lr_example}.  We see
that the single-scale LR approach is moderately faster than block
Gauss-Seidel, but introducing coarser-scales into the method leads to
a significant speed-up in the rate of convergence.

\section{Discussion}

We have introduced a general Lagrangian relaxation framework for MAP
estimation in both discrete and Gaussian graphical models.  This
provides a new interpretation of some existing methods, provides
deeper insights into those methods, and leads to new generalizations,
such as the multiscale relaxation introduced here. There are many promising
directions for further work.  While we have considered discrete and
Gaussian models separately, the basic approach should extend to the
richer class of conditionally Gaussian models \cite{Lauritzen96}
including both discrete and continuous variables.  In discrete models,
designing augmented models that capture more structure of the original
problem leads to reduced duality gaps and optimal MAP estimates in
larger classes of models.  It would be of great interest to finds ways
to \emph{adaptively} search this hierarchy of relaxations to
efficiently reduce and eventually eliminate the duality gap with
minimal computation.  It is also of interest to consider
approaches to identity provably \emph{near-optimal} estimates, perhaps
using the relaxed max-marginal estimates, in cases where it is not
tractable to completely eliminate the duality gap.


\bibliography{lr}
\bibliographystyle{unsrt}

\end{document}